\documentclass[sigplan,screen]{acmart}
\AtBeginDocument{%
  }

\setcopyright{acmlicensed}
\copyrightyear{2018}
\acmYear{2018}
\acmDOI{XXXXXXX.XXXXXXX}
\acmConference[Conference acronym 'XX]{Make sure to enter the correct
  conference title from your rights confirmation email}{June 03--05,
  2018}{Woodstock, NY}
\acmISBN{978-1-4503-XXXX-X/2018/06}

\usepackage{soul}
\usepackage{url}
\usepackage{hyperref}
\usepackage[utf8]{inputenc}
\usepackage{caption}
\usepackage{graphicx}
\usepackage{amsmath}
\usepackage{amsthm}
\usepackage{pifont}
\usepackage{booktabs}
\usepackage{lscape}
\usepackage{subcaption}
\usepackage{algorithm}
\usepackage{algorithmic}
\usepackage[switch]{lineno}
\usepackage{microtype}

\usepackage{amsmath,amsfonts,bm}

\def\eqref#1{equation~\ref{#1}}

\def\1{\bm{1}}

\DeclareMathAlphabet{\mathsfit}{\encodingdefault}{\sfdefault}{m}{sl}
\SetMathAlphabet{\mathsfit}{bold}{\encodingdefault}{\sfdefault}{bx}{n}

\usepackage{natbib}
\usepackage{pythonhighlight}
\usepackage{cleveref}

\usepackage{nicefrac}
\usepackage{multirow}
\usepackage{lineno}

\usepackage{tabularx}
\usepackage{xspace}
\usepackage{comment}
\usepackage{enumitem}

\usepackage[nolist]{acronym}
\begin{acronym}[UML]
	\acro{KG}{Knowledge Graph}
    \acro{KGE}{Knowledge Graph Embedding}
    \acro{FOL}{First-Order Logic}
    \acro{AI}{Artificial Intelligence}
    \acro{EPFO}{Existential Positive First-order}
    \acro{DAG}{Directed Acyclic Graph}
    \acro{DNF}{Disjunctive Normal Form}
	\acro{AI}{Artificial Intelligence}
	\acro{CL}{Concept Learning}
	\acro{CEL}{Class Expression Learning}
	\acro{CWA}{Close World Assumption}
	\acro{CSV}{Comma-Separated Values}
	\acro{CNN}{Convolutional Neural Network}
	\acro{DL}{Description Logic}
	\acro{DNN}{Deep Neural Network}
    \acro{FOL}{First-Order Logic}
  	\acro{GD}{Gradient Descent}
    \acro{KB}{Knowledge Base}
    \acro{KG}{Knowledge Graph}
    \acro{KGE}{Knowledge Graph Embedding}
	\acro{SW}{Semantic Web}
	\acro{OWA}{Open World Assumption}
	\acro{SGD}{Stochastic Gradient Descent}
	\acro{KGC}{Knowledge Graph Completion}

	\acro{MAP}{Maximum a Posteriori Probability Estimation}
	\acro{MRR}{Mean Reciprocal Rank}
	\acro{ML}{Machine Learning}
    \acro{MLE}{Maximum Likelihood Estimation}
    \acro{MDP}{Maximum Likelihood Estimation}
	\acro{MTL}{Multi-task Learning}
	\acro{MDP}{Markov Decision Process}
	\acro{NN}{Neural Network}

	\acro{KP}{Kronecker Product}
    \acro{KD}{Kronecker Decomposition}
    \acro{KB}{Knowledge Base}
    \acro{KG}{Knowledge Graph}
    \acro{KGE}{Knowledge Graph Embedding}
	\acro{ILP}{Inductive Logic Programming}
	\acro{RL}{Reinforcement Learning}
	\acro{RNN}{Recurrent Neural Network}
	\acro{RDF}{Resource Description Framework}
 	\acro{SWA}{Stochastic Weight Averaging}
	\acro{OWL}{Web Ontology Language}
	\acro{OWA}{Open World Assumption}
    \acro{WWW}{World Wide Web}
\end{acronym}  

\newcommand{\entities}{\ensuremath{\mathcal{E}}\xspace}
\newcommand{\relations}{\ensuremath{\mathcal{R}}\xspace}

\newcommand{\scoreFunc}{\phi}

\newcommand{\approach}{\textsc{Ebr}\xspace}

\copyrightyear{2025}
\acmYear{2025}
\setcopyright{cc}
\setcctype{by}
\acmConference[K-CAP '25]{Knowledge Capture Conference 2025}{December 10--12, 2025}{Dayton, OH, USA}
\acmBooktitle{Knowledge Capture Conference 2025 (K-CAP '25), December 10--12, 2025, Dayton, OH, USA}
\acmDOI{10.1145/3731443.3771348}
\acmISBN{979-8-4007-1867-0/2025/12}

\begin{document}

\title{Neural Reasoning for Robust Instance Retrieval in \texorpdfstring{$\mathcal{SHOIQ}$}{SHOIQ}}

\author{Louis Mozart Kamdem Teyou}
\orcid{0000-0001-7975-8794}
\affiliation{%
  \institution{Heinz Nixdorf Institute, Paderborn University}
  \city{Paderborn}
  \state{NRW}
  \country{Germany}
}

\author{Luke Friedrichs}
\orcid{0009-0000-0883-8316}
\affiliation{%
  \institution{Heinz Nixdorf Institute, Paderborn University}
  \city{Paderborn}
  \state{NRW}
  \country{Germany}
}

\author{N'Dah Jean Kouagou}
\orcid{0000-0002-4217-897X}
\affiliation{%
  \institution{Heinz Nixdorf Institute, Paderborn University}
  \city{Paderborn}
  \state{NRW}
  \country{Germany}
}

\author{Caglar Demir}
\orcid{0000-0001-8970-3850}
\affiliation{%
  \institution{Heinz Nixdorf Institute, Paderborn University}
  \city{Paderborn}
  \state{NRW}
  \country{Germany}
}

\author{Yasir Mahmood}
\orcid{0000-0002-5651-5391}
\affiliation{%
  \institution{Heinz Nixdorf Institute, Paderborn University}
  \city{Paderborn}
  \state{NRW}
  \country{Germany}
}

\author{Stefan Heindorf}
\orcid{0000-0002-4525-6865}
\affiliation{%
  \institution{Heinz Nixdorf Institute, Paderborn University}
  \city{Paderborn}
  \state{NRW}
  \country{Germany}
}

\author{Axel-Cyrille Ngonga Ngomo}
\orcid{0000-0001-7112-3516}
\affiliation{%
  \institution{Heinz Nixdorf Institute, Paderborn University}
  \city{Paderborn}
  \state{NRW}
  \country{Germany}
}

\renewcommand{\shortauthors}{Kamdem et al.}

\begin{abstract}
Concept learning exploits background knowledge in the form of description logic axioms to learn explainable classification models from knowledge bases. Despite recent breakthroughs in neuro-symbolic concept learning, most approaches still cannot be deployed on real-world knowledge bases. This is due to their use of description logic reasoners, which are not robust against inconsistencies nor erroneous data. We address this challenge by presenting a novel neural reasoner dubbed \approach. Our reasoner relies on embeddings to approximate the results of a symbolic reasoner. We show that \approach solely requires retrieving instances for atomic concepts and existential restrictions to retrieve or approximate the set of instances of any concept in the description logic $\mathcal{SHOIQ}$. In our experiments, we compare \approach with state-of-the-art reasoners. Our results suggest that \approach is robust against missing and erroneous data in contrast to existing reasoners.
\end{abstract}

\maketitle

\ccsdesc[300]{Computing methodologies~Machine learning approaches}
\ccsdesc[500]{Computing methodologies~Description logics}
\ccsdesc[500]{Computing methodologies~Ontology engineering}

\keywords{Instance retrieval, Neural reasoner, Symbolic reasoner, Description logic, Concept learning, Knowledge base}



\section{Introduction}

\acp{DL}~\cite{baader2003description} offer a formal framework for structured knowledge representation and reasoning. 
Due to their well-defined semantics and favorable computational properties, \acp{DL} are now essential in fields such as ontology engineering~\cite{keet2018introduction}, knowledge representation~\cite{brachman2004knowledge}, and the Semantic Web~\cite{horrocks2003shiq}. Concept learning over \ac{DL} \acp{KB} is a form of ante-hoc explainable supervised machine learning that returns a \ac{DL} 
concept as a classifier. 

Currently, concept learners rely on symbolic reasoners (e.g., Pellet~\cite{sirin2007pellet}, Fact++~\cite{tsarkov2006fact++} and HermiT~\cite{glimm2014hermit} for instance retrieval. 
Although such reasoners are being successfully applied to infer concept hierarchies in ontological data~\cite{sattler2014does,werner2014using},
their application on real data is limited by their inability to handle inconsistencies and to infer missing instance assertions. An inconsistency occurs when a knowledge base entails a contradiction, that is, when some assertions cannot simultaneously be true given the ontology's axioms.
For example, consider the inconsistent knowledge base $\mathcal{K}=(\{C\sqsubseteq A, C\sqsubseteq B, A\sqcap B \sqsubseteq \bot\}, \{C(a)\})$.
A classical symbolic reasoner cannot determine the class membership of the individual $a$ since it is inferred to be in both $A$ and $B$, which leads to an inconsistency with the axiom $A\sqcap B \sqsubseteq \bot$. By contrast, an incomplete knowledge base lacks certain facts or assertions that are true in the domain but absent from the data.
For instance, $\mathcal{K}'=(\emptyset, 
\{\texttt{Person}(Bob), \texttt{Person}(Paul) , \texttt{Person}(Ani),\\
\texttt{knows}(Bob, Paul), \texttt{knows}(Ani, Joe)\}),$  where class memberships are missing for some individuals, e.g., $Joe$. In $\mathcal K'$,
a symbolic reasoner cannot infer that $Joe$ is an instance of $\texttt{Person}$.

Inconsistency and incompleteness are well-recognized challenges in practical knowledge base applications, as real-world \acp{KB}-particularly large, evolving, or automatically constructed ones are rarely complete or fully consistent~\cite{nickel2015review}. The presence of such imperfections is not only common but also well-documented; for instance, the DBpedia\footnote{https://dbpedia.org/ontology/} ontology contains conflicting assertions such as listing both $\texttt{Virginia}$ and $\texttt{British America}$ as George Washington's birthplace, despite their historical overlap. While standard DL reasoners are capable of detecting inconsistencies, they do not operate reliably in their presence, often halting entirely when even a few conflicting assertions exist in otherwise valid KBs. Our work builds on this motivation by introducing a novel application of knowledge graph embedding models for reasoning under inconsistency and incompleteness, offering a data-driven alternative to symbolic reasoning in settings where classical approaches break down.

With this work, we aim to further the development of scalable and robust concept retrieval by presenting a neural reasoner dubbed \approach (\textbf{E}mbedding \textbf{B}ased \textbf{R}easoner) that is robust against inconsistencies and missing assertions. Neural link predictors have been extensively investigated to deal with incompleteness on various datasets
~\cite{dettmers2018convolutional,ren2020beta}. Recent works showed that neural link predictors can be effectively applied to answer complex queries involving multi-model reasoning~\cite{bai2023answering,demir2023litcqd}. Our approach \approach leverages knowledge graph embeddings to perform robust instance retrieval over incomplete and inconsistent KBs. It relies on a simple neural link predictor to implement the robust retrieval of assertions from incomplete and inconsistent data. These assertions are combined via set-theoretic operations to compute instances of concepts. A summary of our contributions is as follows:
\begin{itemize}[noitemsep, topsep=3.5pt]
 \item We propose \approach, a neural semantics that leverages knowledge graph embeddings to perform concept learning and instance retrieval over expressive \acp{KB} formulated in $\mathcal{SHOIQ}$, thereby extending the use of embedding models beyond traditional link prediction tasks.
  \item We provide an in-depth comparison of \approach against symbolic reasoners on six datasets. We show that our approach outperforms symbolic methods on knowledge bases with a varying percentage of false or missing assertions.
 \item The source code of \approach, along with the datasets and scripts used to reproduce these experiments, is publicly available\footnote{https://github.com/Louis-Mozart/Embedding-Based-Reasoner} to support further development.
\end{itemize}

\section{Background and Related Works}

\subsection{Knowledge Bases in Description Logics}
A knowledge base (\ac{KB}) is a pair $\mathcal{K} = \langle \mathcal T, \mathcal A\rangle$.
The TBox $\mathcal T$ contains general concept inclusions (GCIs) of the form ${C}  \sqsubseteq {D}$, where ${C}, {D}$ are concepts.
The ABox $\mathcal A$ includes assertions having the form ${C}(a)$ (concept assertion) or ${r}(a, b)$ (role assertion) for individuals $a, b$, concepts ${C}$, and role ${r}$.
The description logic $\mathcal{SHOIQ}$ additionally includes an RBox, which contains axioms pertaining to roles.
A \emph{role inclusion axiom} (RIA) has the form $r \sqsubseteq s$. A \emph{transitivity axiom} is denoted by $\text{Tra}(r)$, and a \emph{functionality axiom} by $\text{Fun}(r)$, where $r$ and $s$ are roles.
A role $s$ is simple if there is neither a transitivity axiom $\text{Tra}(r)$, nor $\text{Tra}(r^{-1})$ for any role $r$ subsumed by $s$.
For simplicity of presentation, we assume that the TBox $\mathcal T$ includes both concept and role axioms.
The syntax and semantics for concepts in $\mathcal{SHOIQ}$~\cite{baader2003description,hitzler2009foundations} are given in Table~\ref{tab:semantics1}.

\begin{table}[htb]
	\centering
    \caption{Syntax and semantics for $\mathcal{SHOIQ}$ concepts. The role $s$ in cardinality restriction is a simple role.  
$\mathcal{I}$ stands for an interpretation with domain $\Delta^\mathcal{I}$. 
 } 
 \resizebox{\columnwidth}{!}{%
   \begin{tabular}{@{}lccc@{}}
        \toprule
		\textbf{Construct}               & \textbf{Syntax}         & \textbf{Semantics} \\
		\midrule
        Atomic concept          & $A$            & $A^{\mathcal{I}}\subseteq{\Delta^\mathcal{I}}$ \\
        Atomic role                    & $r$            & $r^\mathcal{I}\subseteq{\Delta^\mathcal{I}\times \Delta^\mathcal{I}}$\\
  		Top concept             & $\top$         & $\Delta^\mathcal{I}$\\
  		Bottom concept          & $\bot$         & $\emptyset$            \\
 		
 		Negation                & $\neg C$       & $\Delta^\mathcal{I}\setminus C^\mathcal{I}$
 		\\
 		Conjunction             & $C\sqcap D$    & $C^\mathcal{I}\cap D^\mathcal{I}$ \\
 		Disjunction             & $C\sqcup D$    & $C^\mathcal{I}\cup D^\mathcal{I}$\\
 		Existential restriction & $\exists\  r.C$ & $\{ x \mid \exists\ y. (x,y) \in 
   r^\mathcal{I} \land y \in C^\mathcal{I}\}$\\
        Inverse Role & $r^{-1}$ & $\{(y,x) \mid (x,y)\in r^\mathcal{I}\}$\\
        Nominals & $\{o\}$ & $\{o\}^{\mathcal{I}}\subseteq \Delta^{\mathcal{I}}$ \\ 
 		 At least restriction & $\geq n s.C$  & $\{a \mid \; |  \{b\in C | (a,b) \in s^\mathcal{I} \}| \geq n \}$\\
 		 At most restriction &$\leq n s.C$  & $\{a \mid \; |  \{b\in C| (a,b) \in s^\mathcal{I} \}| \leq n \}$\\
		\bottomrule
	\end{tabular}}

 \label{tab:semantics1}
\end{table}
Let $C\sqsubseteq D$ be a GCI and $\mathcal I$ be an interpretation.
Then, $\mathcal I$ satisfies $C\sqsubseteq D$, denoted as $\mathcal I\models {C}\sqsubseteq D$ iff ${C}^\mathcal I\subseteq {D}^\mathcal I$.
Similarly, $\mathcal I$ satisfies an assertion ${C}(a)$ iff $a^\mathcal I\in {C}^\mathcal{I}$. The assertion ${r}(a, b)$ is satisfied by $\mathcal I$ iff $(a^\mathcal I , b^\mathcal I ) \in {r}^\mathcal I$.
Finally, $\mathcal I$ satisfies an RIA $r \sqsubseteq s$ iff $r^{\mathcal I}\subseteq s^{\mathcal I}$ and axioms $\text{Tra}(r),\text{Fun}(r)$ iff the interpretation $r^{\mathcal I}$ is a transitive and a functional role, respectively.
We say that $\mathcal I$ is a model of the KB $\mathcal K$, denoted by $\mathcal I \models \mathcal K$, iff $\mathcal I$ satisfies every axiom in $\mathcal K$.
Finally, let $\mathcal K$ be a DL KB and $\alpha$ be an axiom, then $\mathcal K\models \alpha$ iff $\mathcal I\models\alpha$ for every model $\mathcal I$ of $\mathcal K$.


\subsection{Symbolic Reasoning over Knowledge Bases}
Reasoning with expressive \ac{DL} \acp{KB} is a computationally hard task. Specifically, the instance checking problem, which, given~$\mathcal{K}$, a concept~$C$, and an individual $x$, determines whether $x$ is an instance of $C$ in $\mathcal{K}$ (denoted as $\mathcal{K} \models C(x)$) 
is non-deterministic (resp., double) exponential time complete for $\mathcal{SHOIQ}$~\cite{Kazakov08}.
Given such high complexity and the additional challenges posed by incomplete and inconsistent data in real-world scenarios, practical applications often require the use of approximation algorithms.

To the best of our knowledge, HermiT~\cite{glimm2014hermit} is the only reasoner that fully supports the OWL 2 standard. 
HermiT is based on a 
``hypertableau'' calculus that addresses performance problems due to nondeterminism and model size---the primary sources of complexity in state-of-the-art OWL reasoners. 
HermiT was shown to outperform the previous reasoners, including Pellet~\cite{sirin2007pellet}, and Fact++~\cite{tsarkov2006fact++}. OWL2Bench~\cite{Singh2020OWL2Bench} compares different reasoners across datasets and OWL profiles. Pellet and its extension Openllet were found to perform best in terms of runtime. JFact, the Java implementation of Fact++, performed worst on all the reasoning tasks across all OWL~2 profiles. While OWL2Bench assessed the reasoner's performance to detect inconsistent ontologies, they did not assess the robustness of reasoners, i.e., how well they answer queries on incomplete or inconsistent data.

\subsection{Knowledge Graph Embeddings}
\label{subsec:kge}
Let $\mathcal{E}$ be the set of all entities and  $\mathcal{R}$ the set of all relations.
A \emph{knowledge graph} (KG) $\mathcal K \subseteq\entities \times \relations \times \entities $ is a set of triples (also called \emph{assertions}) of the form  $(x, r, y)$  where $x, y\in \mathcal E$ are called head and tail entities, respectively, and $r \in \mathcal R$ depict the relationship between them. 
A \ac{KGE} model is defined as a parameterized scoring function over KGs, depicted as $\scoreFunc_\Theta: \mathcal{K}\rightarrow \mathbb{R}$.
where $\Theta$ denotes parameters, which often comprise 
entity embeddings $\mathbf{E} \in \mathbb{R}^{|\mathcal{E}| \times d_e}$, relation embeddings $\mathbf{R} \in \mathbb{R}^{|\mathcal{R}| \times d_r}$, and additional parameters (e.g., affine transformations, batch normalizations, convolutions).
Since $d_e = d_r$ holds for many state-of-the-art models, we will use $d$ to denote the number of embedding dimensions for both entities and relation types in a knowledge graph.

A plethora of \ac{KGE} models have been developed over the last decade to address reasoning tasks over large symbolic knowledge graphs~\cite{dai2020survey,wang2021survey}. These models aim to map entities and relations from the KG  into continuous vector spaces, enabling efficient and scalable computation via algebraic operations. Most \ac{KGE} approaches are primarily designed for the link prediction task, where the goal is to estimate the plausibility of a triple $(x, r, y) \in \entities \times \relations \times \entities$. This is typically done via a scoring function $\scoreFunc_\Theta(x, r, y)$ parameterized by $\Theta$, which produces a real-valued score $\hat{y}$ that reflects the likelihood (not necessarily calibrated) of the triple being valid~\cite{demir2023clifford}.

Scoring functions vary across models: translational models like TransE~\cite{bordes2013translating} interpret relations as translations in vector space (i.e., $\vec{x} + \vec{r} \approx \vec{y}$), while bilinear models like DistMult~\cite{yang2014embedding} and ComplEx~\cite{trouillon2017knowledge} employ multiplicative interactions, capturing symmetric and asymmetric patterns. Recent work has introduced more expressive representations based on Clifford algebras. Notably, Keci~\cite{demir2023clifford} generalizes models like DistMult and ComplEx by embedding entities and relations in Clifford spaces of the form $Cl_{p,q}$. DeCaL~\cite{teyou2024embedding} extends this framework using degenerate Clifford algebras $Cl_{p,q,r}$, achieving state-of-the-art results on standard benchmarks.

Since most knowledge graphs encode only positive (i.e., true) facts, training \ac{KGE} models necessitates the creation of negative samples typically via corrupting the head or tail entity in a triple. This is done using negative sampling techniques such as Bernoulli sampling \cite{qian2021understanding}, 1-vs-All, or K-vs-All strategies~\cite{Rufinelli2020You}, which help the model differentiate between plausible and implausible triples. Learning is commonly performed using margin-based ranking losses or cross-entropy objectives, with regularization techniques to mitigate overfitting in large-scale settings.

\acp{KGE} have demonstrated remarkable versatility and effectiveness across a wide range of applications beyond link prediction. In drug discovery, embeddings assist in identifying novel drug-target interactions by modeling complex biomedical graphs~\cite{bonner2022understanding}. In question answering, KGEs serve as soft reasoning engines that embed semantic relationships into the retrieval and inference pipeline~\cite{hamilton2018embedding}. In recommender systems, they enhance traditional collaborative filtering by leveraging multi-relational user-item graphs~\cite{choudhary2021self}. Moreover, recent trends explore explainable embeddings, temporal dynamics, and inductive generalization, pushing KGE techniques closer to deployment in real-world intelligent systems.

\section{Approach}
 
\approach by itself is not an embedding model, but a reasoning framework that uses the representations learned by an embedding model to perform reasoning tasks like instance retrieval or concept learning. While KGE models focus on learning vector representations for entities and relations, EBR builds on top of these representations to approximate logical inference.

\subsection{Knowledge Graph Embedding and Construction}
Mapping description logic syntax to our neural semantics (see in Section~\ref{neural-semantics}) solely requires an engine that can accurately predict whether an assertion $(x, r, y)$ is true. We map this task to computing whether the score assigned to an assertion is greater than or equal to a preset threshold $\gamma$. 
A key to achieving high accuracy on this task lies in the construction of a knowledge graph that renders the key assertions for our input knowledge base in such a way that accurate predictions can be carried out. To construct this knowledge graph $\mathcal{K}\subseteq \mathcal{E} \times \mathcal{R}\times \mathcal{E}$, we extract taxonomic axioms of the form $C(a)\equiv (x, \texttt{rdf:type}, C),\ r(x,y)\equiv (x,r,y)$, $C\sqsubseteq D\equiv (C, \texttt{rdfs:subClassOf}, D)$ and $r \sqsubseteq s\equiv (r, \texttt{rdfs:subPropertyOf}, s)$. The graph $\mathcal G$ is then used as input to a KGE model to learn embeddings for entities and relation types.
This yields a trained KGE model $\phi_\Theta: \mathcal{E} \times \mathcal{R}\times \mathcal{E} \rightarrow \mathbb{V}^d$--- where $\mathbb{V}^d$ is a vector space.
\vspace{-0.01cm}
\begin{align}
&\phi_\Theta: \mathcal{E} \times \mathcal{R}\times \mathcal{E} \rightarrow \mathbb{R};
&\phi_\Theta(x, r, y) = {Re}(\langle\mathbf{x},\mathbf{r},\bar{\mathbf{y}}\rangle).\end{align}
Here, $\mathbf{x}$, $\mathbf{r}$, and $\mathbf{y}$ are the complex embeddings of the head entity $x$, the relation type $r$, and the tail entity $y$, respectively. ${Re}$ stands for the real part of a complex number. Finally, $\overline{\mathbf{y}}$ denotes the complex conjugate of $\mathbf{y}$ and $\langle \mathbf{x},\mathbf{r}, \mathbf{y}\rangle$ stands for the trilinear product of $\mathbf{x}$, $\mathbf{r}$ and $\mathbf{y}$.

\subsection{Prediction Mechanism}
Once the model $\phi_\Theta$ is trained, we construct a neural link predictor $\phi: \mathcal E \times \mathcal R \times \mathcal E \rightarrow [0,1]$ which can assign a truth probability to any triple $(x,r,y)$ from its domain. Given that the scores computed by $\phi_\Theta$ are not bound, we chose to use the sigmoid trick used in neural networks to translate scores into a range that can be interpreted as a probability. Consequently, we set $\phi(x, r, y) = \sigma(\phi_\Theta(x, r, y))$, where $\sigma$ stands for the sigmoid function. A triple is then considered true in our neural interpretation if $\phi(x, r,y)$ is larger than a preset threshold $\gamma$. This is a key difference to symbolic reasoning (see Table \ref{tab:neuralsemantics} for details on the neural semantics). In particular, triples $(x, r, y)$ found in $\mathcal{K}$ might be predicted to be false and hence not included in the neural interpretation of the input knowledge graph.

\subsection{Mapping DL Syntax to Neural Semantics}
\label{neural-semantics}

The syntax and semantics for concepts in $\mathcal{SHOIQ}$ are provided in Table \ref{tab:semantics1}. 
We define a mapping from concept constructors in $\mathcal{SHOIQ}$ to their neural semantics in Table \ref{tab:neuralsemantics}. For a given \ac{KB} $ \mathcal K$, the domain $\Delta^\mathcal N \subseteq \mathcal{E} \cup \mathcal{R}$ of the interpretation $\mathcal{N}$ computed by \approach is the set of all class names, individual names and role names found in the axioms in $\mathcal K$. 

\begin{table}[htb]
	\centering
    \caption{Syntax \& neural semantics for $\mathcal{SHOIQ}$ concepts. 
$\Delta^\mathcal{N}$ stands for the domain of the neural interpretation $\mathcal{N}$ of $\approach$. $s$ stands for a simple role. 
 } 
\resizebox{\columnwidth}{!}{%
   \begin{tabular}{@{}lccc@{}}
        \toprule
		\textbf{Construct}               & \textbf{Syntax}         & \textbf{Neural Semantics $\cdot^{\mathcal N}$}\\
		\midrule
        Atomic concept          & $A$            & $\{x \mid  \phi(x,\texttt{rdf:type},A) \geq \gamma\}$ \\
        Atomic role                    & $r$            & $\{(x,y) \mid  \phi(x,r,y) \geq \gamma\}$\\
  		Top concept             & $\top$         & $\Delta^\mathcal{N}$\\
  		Bottom concept          & $\bot$         & $\emptyset$            \\
 		
 		Negation                & $\neg C$       & $\Delta^\mathcal{N}\setminus C^{\mathcal N}$
 		\\
 		Conjunction             & $C\sqcap D$    & $C^{\mathcal N} \cap D^{\mathcal N}$ \\
 		Disjunction             & $C\sqcup D$    & $C^{\mathcal N} \cup D^{\mathcal N}$\\
 		Existential restriction & $\exists\  r.C$ & $\{ x \mid \exists\ y \text{ s.t. } (x,y) \in r^{\mathcal N} \land y \in C^{\mathcal N}\}$\\
 		Universal restriction & $\forall\   r.C$   & $(\neg (\exists r. \neg C))^{\mathcal N}$
\\ 	
        Universal Role                    & $U$            & $\Delta^\mathcal{N}\times \Delta^\mathcal{N}$\\        
        Inverse Role & $r^{-1}$ & $\{(y,x) \mid (x,y)\in r^{\mathcal N} \}$\\
        Nominals & $\{o\}$ & $\{o^{\mathcal N}\}$ \\ 
 		 At least restriction & $\geq n s.C$  & $\{a \mid \; |  \{b\in C^{\mathcal N} | (a,b) \in s^{\mathcal N} \}| \geq n \}$\\
 		 At most restriction &$\leq n s.C$  & $\{a \mid \; |  \{b\in C^{\mathcal N}| (a,b) \in s^{\mathcal N} \}| \leq n \}$\\
		\bottomrule
	\end{tabular}}

 \label{tab:neuralsemantics}
\end{table}

These mappings allow us to handle not only atomic concepts and roles but also more expressive axioms involving complex concept expressions. For example, consider the TBox axiom $A \sqcap \exists r. B \sqsubseteq C \sqcup \forall r. D$, which asserts that individuals belonging to both $A$ and those related via $r$ to instances of $B$ must also belong to either $C$ or to those whose $r$-successors are all in $D$. In the neural setting, this axiom is interpreted as a set inclusion constraint:
$(A^{\mathcal{N}} \cap \{ x \mid \exists y.\ (x, y) \in r^{\mathcal{N}} \land y \in B^{\mathcal{N}} \}) \subseteq (C^{\mathcal{N}} \cup \{ x \mid \forall y.\ (x, y) \in r^{\mathcal{N}} \Rightarrow y \in D^{\mathcal{N}} \})$.

While such axioms are not explicitly encoded as triples in the knowledge graph, their components, such as class memberships and role assertions, are grounded via basic triples like \texttt{:a rdf:type :A}, \texttt{:a :r :b}, and \texttt{:b rdf:type :B}. The embedding-based interpretation $\mathcal{N}$ reconstructs the semantics of the full axiom by composing the neural representations of these grounded components, thus enabling reasoning over complex concept expressions in a continuous space.

\section{Experimental Setup}
\label{sec:Experimental Setup}
\approach is designed to improve the robustness of concept learning. In our experiments, we evaluate its performance on instance retrieval under varying levels of incompleteness and inconsistency, and further demonstrate that its reasoning capabilities generalize to concept learning as well.

Candidate concepts are generated using the refinement operator $\rho$ from \cite{demir2023neuro} by randomly sampling concepts from its refinement tree. The next section presents the datasets and evaluation metrics, followed by the experimental results and analysis.

\subsection{Datasets}
\label{sec:datasets}
We evaluated our proposed approach on six benchmark datasets, including four large datasets (\textit{Carcinogenesis}, \textit{Mutagenesis}, \textit{Semantic Bible}, and \textit{Vicodi}) and two smaller datasets (\textit{Family} and \textit{Father}). These datasets cover a range of domains, from biological interactions to historical and familial relationships. Detailed statistics for each dataset are provided in Table~\ref{tab:datasets}. 

\begin{table}[tb]
\centering
\setlength{\tabcolsep}{2.3pt}
\caption{Detailed information about the datasets used for evaluation}
\begin{tabular}{@{}lcccccc@{}}
    \toprule
    \textbf{Dataset} & $|$\textbf{Ind.}$|$ & $|$\textbf{Classes}$|$ & $|$\textbf{Prop.}$|$ & $|\textbf{TBox}|$ & $|\textbf{ABox}|$\\
    \midrule
    Carcinogenesis & 22,372 & 142 &\phantom{0}4 & 144 &\phantom{0}74,223 \\
    Mutagenesis & 14,145 &\phantom{0}86 &\phantom{0}5 & \phantom{0}82 &\phantom{0}47,722 \\
    Semantic Bible &\phantom{000}724 &\phantom{0}48 & 29 &\phantom{0}56 &\phantom{00}3,106 \\
    Vicodi & 33,238 & 194 & 10 & 204 & 116,181 \\
    Family & \phantom{00}202 & \phantom{0}18 & \phantom{0}4 & \phantom{0}26 & \phantom{000}472 \\
Father & \phantom{0000}6 & \phantom{00}4 & \phantom{0}1 & \phantom{0}3 & \phantom{00000}4 \\
    \bottomrule
\end{tabular}

\label{tab:datasets}
\end{table}

\subsection{Tasks}

We evaluate \approach across three main tasks to assess its robustness and effectiveness under different knowledge base conditions.

\paragraph{Task 1: Instance Retrieval.}
In the first task, we evaluate \approach on standard instance retrieval using complete and consistent knowledge bases. 
The goal is to measure how accurately our reasoner retrieves instances of given class expressions compared to symbolic reasoners.
We quantify performance using the Jaccard similarity, which compares the set of instances $\hat{y}$ retrieved by our reasoner with the ground-truth set $y$ obtained from symbolic reasoners.\footnote{Note that all symbolic reasoners used in this setting yield identical results under consistent and complete data.}  
The Jaccard similarity $J(\hat{y}, y)$ is defined as:
\begin{align}
J(\hat{y}, y) = 
\begin{cases}
\dfrac{|\hat{y} \cap y|}{|\hat{y} \cup y|}, & \text{if } y \neq \emptyset \text{ or } \hat{y} \neq \emptyset, \\
1, & \text{otherwise.}
\end{cases}
\end{align}

\paragraph{Task 2: Robustness to Noise and Incompleteness.}
The second task examines the performance of \approach under noisy and incomplete knowledge bases.  
Starting from a consistent and correct knowledge base $\mathcal{K}$, we simulate noise by injecting $\nu|\mathcal{K}|$ random false axioms, where $\nu$ denotes the predefined noise ratio. To simulate incompleteness, we randomly remove $\nu|\mathcal{K}|$ axioms from $\mathcal{K}$. For concept learning, we focus on the inconsistent scenario and compare \approach, against baseline reasoners.

\paragraph{Evaluation Protocol.}
We first choose a KGE model and integrate it into EBR to perform instance retrieval and concept learning across the scenarios described above.  For all instance retrieval experiments, we use the Fast Instance Checker (FIC) reasoner from the DL-Learner framework \cite{JMLR:v26:24-1113,buhmann2016dl}, which can support reasoning under inconsistency, making it a suitable reference for evaluating embedding-based retrieval.
Other symbolic reasoners used include Pellet, HermiT, Openllet, JFact, and the Structural reasoner, the latter being a lightweight variant of FIC.
For concept learning, we evaluate four state-of-the-art learners: OCEL and CELOE \cite{lehmann2010concept,buhmann2016dl} from the DL-Learner framework, CLIP \cite{kouagou2022learning}, and EvoLearner (EVO) \cite{heindorf2022evolearner} with baseline reasoners in the backend and compare their performances (using the traditional F1-score \cite{demir2023neuro}). All concept learners rely on FIC as their default reasoning component, ensuring consistency and comparability across models. Their performances are assessed using the traditional F1-score \cite{demir2023neuro}.

\subsection{Hardware}
All our experiments were carried out on a virtual machine with 2 NVIDIA H100-80C GPUs, each with 80 GB of memory. Note that the embedding computation only consumed approximately 16 GB of GPU memory on the biggest datasets (e.g., \textit{Vicodi}). 



\section{Results and Discussion}

In our experiments, Pellet, HermiT, Openllet, and JFact consistently exhibit identical performance for instance retrieval. Therefore, in this section, we report only the results of Pellet as representative of all four reasoners.
\subsection{KGE Model Selection}
\label{subsec:kge-selection}

\begin{figure}[h!tb]
    \centering
    \subcaptionbox*{}
        {\includegraphics[width=0.23\textwidth, height=2.5cm]{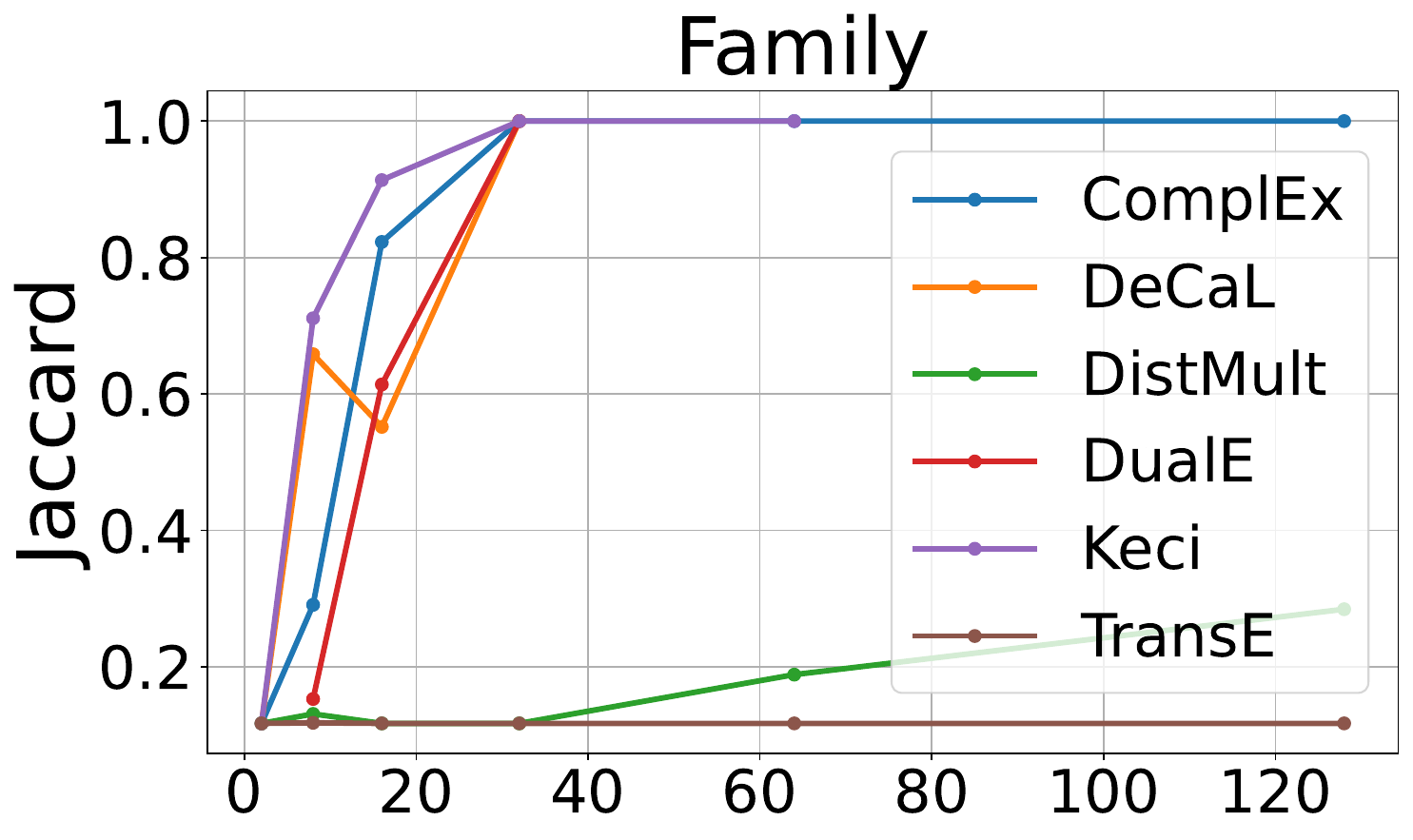}}
        \subcaptionbox*{}{\includegraphics[width=0.23\textwidth, height=2.5cm]{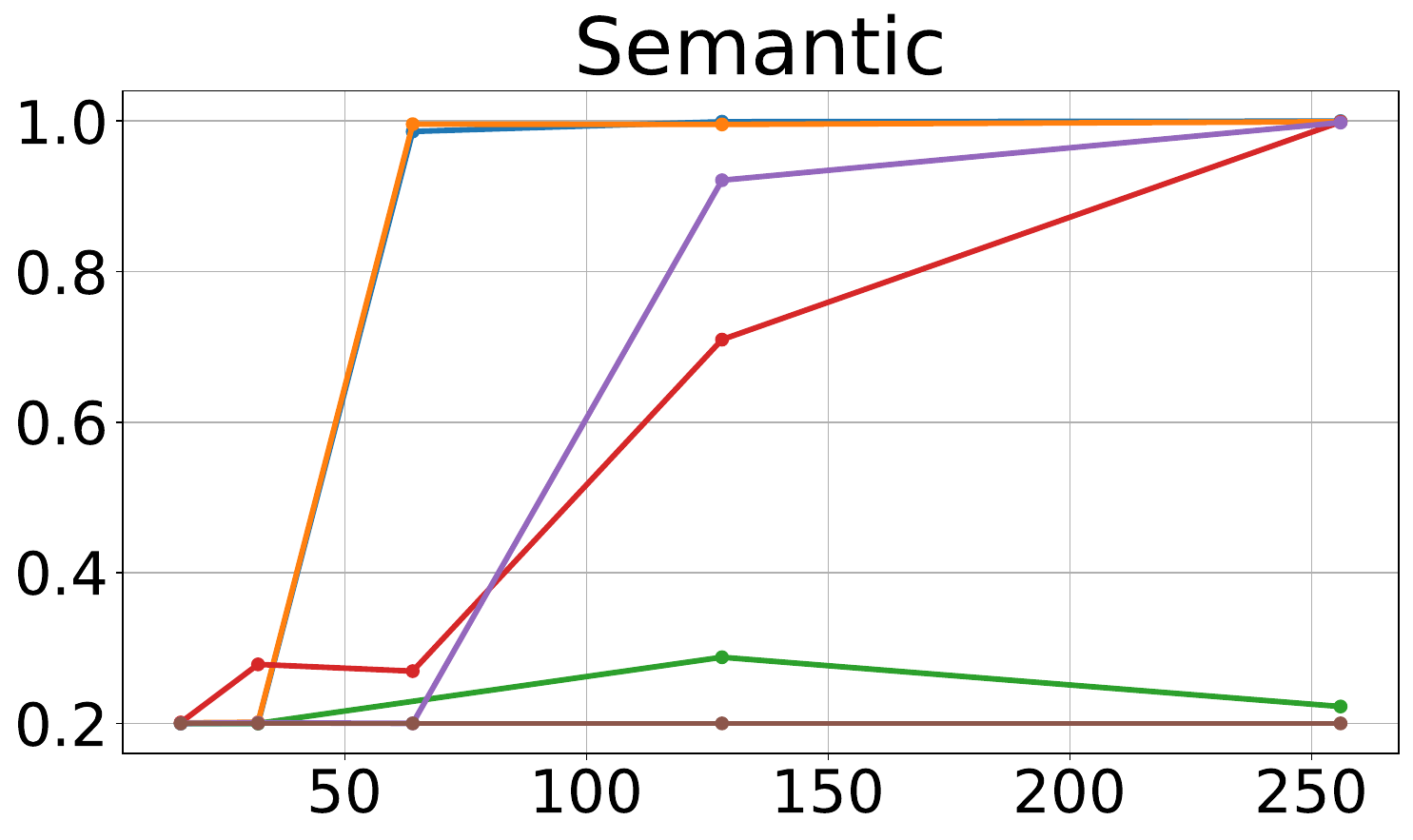}}\\
        \vspace{-.6cm}
        \subcaptionbox*{}{\includegraphics[width=0.23\textwidth, height=2.5cm]{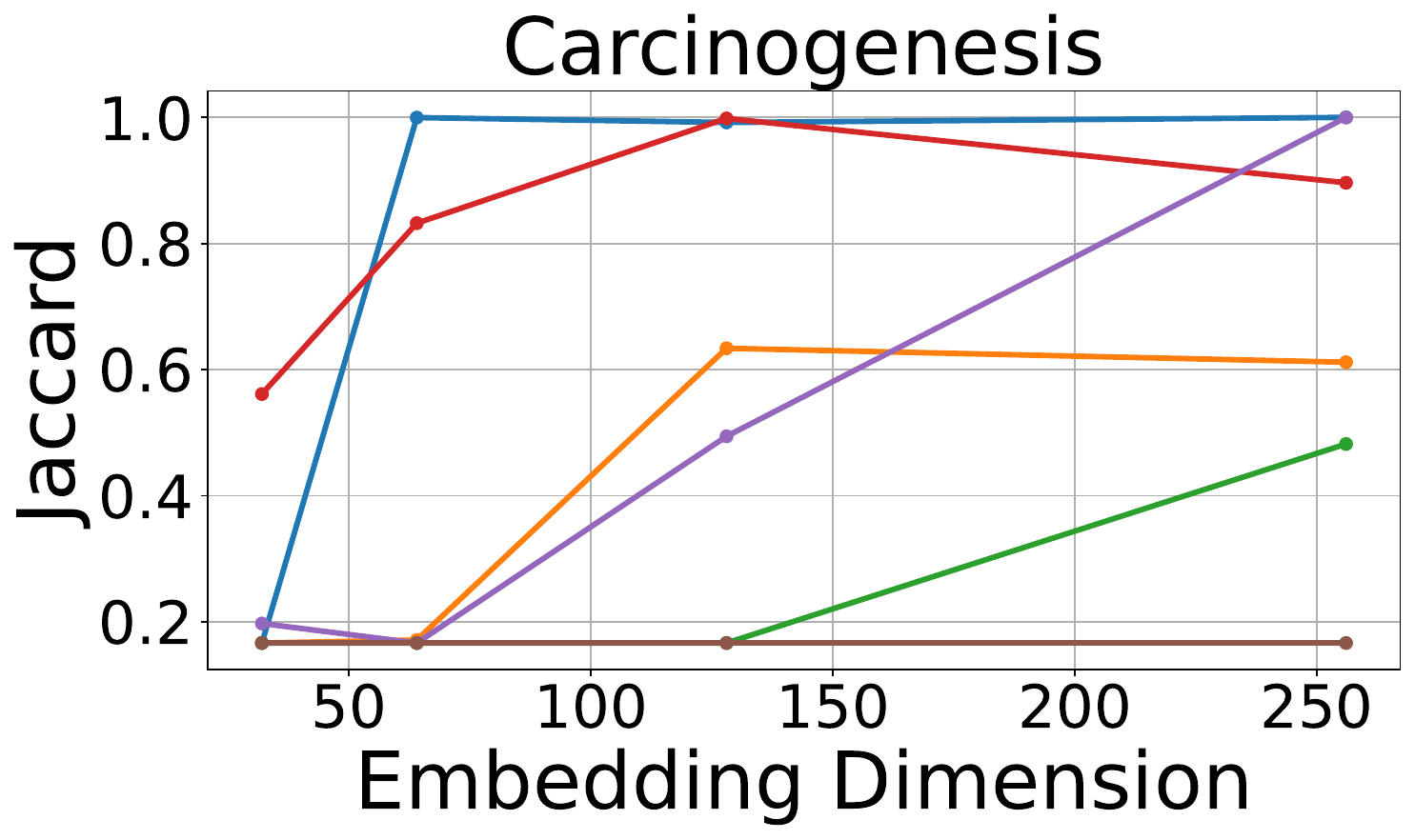}}
        \subcaptionbox*{}{\includegraphics[width=0.23\textwidth, height=2.5cm]{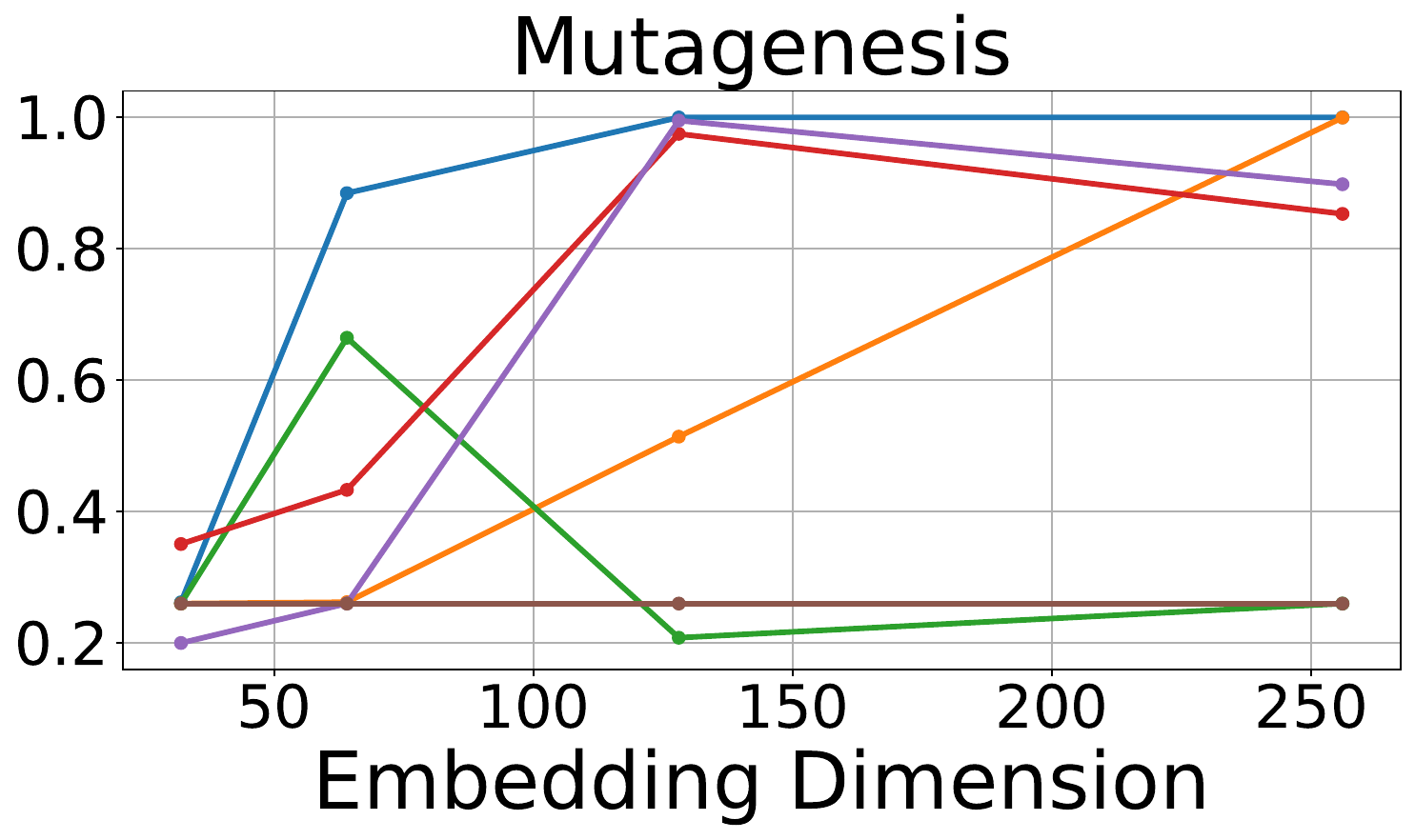}}
        \vspace{-.8cm}
 \caption{\approach average performance with different embedding models and dimensions on different datasets. 
All models share the same legend shown once (right of the top-left panel).}
    \label{fig:jaccard_vs_embedding_datasets}
\end{figure}

We selected the KGE model and embedding dimensionality used throughout our experiments following a systematic protocol aimed at maximizing retrieval accuracy while maintaining a reasonable balance between model complexity and training cost. Each candidate KGE model was trained with embedding dimensions of 16, 32, 64, 128, and 256, and its instance retrieval performance within \approach was assessed using the Jaccard similarity between the instances predicted by the Embedding-Based Reasoner (EBR) and those retrieved by the symbolic instance checker (FIC). The average Jaccard scores over all target expressions are summarized in Figure~\ref{fig:jaccard_vs_embedding_datasets}.

From the figure, it is evident that TransE and DistMult perform poorly--TransE never exceeds a Jaccard score of 0.2, while DistMult remains below 0.4--regardless of the embedding dimensionality. This suggests that the embeddings learned by these models are not sufficiently expressive to capture the relational and hierarchical semantics required for accurate instance retrieval within expressive ontologies. In contrast, other embedding models such as DeCaL, Keci, DualE \cite{cao2021dual}, and ComplEx exhibit strong performance, with their Jaccard similarity steadily converging toward 1.0 as the embedding dimension increases. This indicates that \approach can operate effectively with a wide range of expressive KGE models.

Based on these observations, we selected ComplEx with an embedding dimension of 128 as the default configuration for \approach. This choice was motivated by three consistent findings from Figure~\ref{fig:jaccard_vs_embedding_datasets}: (i) ComplEx achieves the highest and most stable Jaccard scores across all datasets, (ii) performance improvements plateau beyond 128 dimensions, and (iii) the model remains computationally efficient at this size. 
As confirmed in Table~\ref{tab:top_concept_types_datasets}, this configuration enables EBR to achieve perfect retrieval accuracy across all datasets and OWL concept types from dimension 128 onward.

\subsection{Retrieval on Complete and Consistent Datasets}

Table~\ref{tab:top_concept_types_datasets} reports the instance retrieval performance of \approach across four error-free benchmark datasets. Performance improves steadily with larger embedding dimensions, reaching perfect Jaccard scores (1.00) for all concept types at $d=128$ and beyond. The largest gains occur between $d=32$ and $d=64$, indicating that a minimum embedding capacity is needed to capture the ontological structure of these datasets.

For instance, in the Carcinogenesis dataset, accuracy for conjunctions ($C\sqcap D$) rises from 0.37 at $d=32$ to 0.99 at $d=64$, achieving perfection at $d=128$. Similar trends hold for other complex concepts such as $C\sqcup D$, $\exists r.C$, and $\forall r.C$. Increasing $d$ beyond 128 yields no meaningful improvement, confirming that $d=128$ strikes an effective balance between accuracy and computational efficiency. The consistent results across all datasets further support the generality of this configuration.

An important advantage of \approach is its robustness to the choice of the parameter $\gamma$, which governs the selection of individuals (see Table~\ref{tab:neuralsemantics}). Across all datasets, we observed that varying $\gamma$ had a negligible impact on the Jaccard similarity, hence, on the set of retrieved instances. This makes \approach easier to tune in practice, as it consistently retrieves correct instances without requiring fine-grained threshold calibration.

\begin{table}[ht]
\centering
\caption{Average reasoning performance (Jaccard similarity) of \approach using the ComplEx knowledge graph embedding model across different OWL concept types and embedding dimensions on different datasets.}
\resizebox{.48\textwidth}{!}{
\begin{tabular}{@{}lccccccccc@{}}
\toprule
\textbf{Syntax} 
& \multicolumn{4}{c}{\textbf{Carcinogenesis}} 
& \multicolumn{4}{c}{\textbf{Mutagenesis}} \\
\cmidrule(lr){2-5} \cmidrule(lr){6-9}
& \textbf{d=32} & \textbf{d=64} & \textbf{d=128} & \textbf{d=256} 
& \textbf{d=32} & \textbf{d=64} & \textbf{d=128} & \textbf{d=256} \\
\midrule
$A$                & 0.00 & 1.00 & 1.00 & 1.00 & 0.20 & 0.82 & 0.99 & 1.00 \\
$\neg A$              & 0.00 & 0.99 & 1.00 & 1.00 & 0.00 & 0.80 & 0.99 & 1.00 \\
$C\sqcap D$            & 0.37 & 0.99 & 1.00 & 1.00 & 0.55 & 0.90 & 0.99 & 1.00 \\
$C\sqcup D$           & 0.00 & 0.99 & 1.00 & 1.00 &0.06 & 0.90 & 0.99 & 1.00 \\
$\exists r.C$           & 0.00 & 0.99 & 1.00 & 1.00 & 0.00 & 0.78 & 0.99 & 1.00 \\
$\forall r.C$             & 0.21 & 0.98 & 1.00 & 1.00 & 0.00 & 0.75 & 0.99 & 1.00 \\
$\geq n s.C$  & 0.00 & 0.97 & 1.00 & 1.00 & 0.02 & 0.90 & 0.99 & 1.00 \\
$\leq n s.C$   & 0.00 & 0.98 & 1.00 & 1.00 & 0.00 & 0.90 &0.99 & 1.00 \\
$\{o\}$             & 0.00 & 0.97 & 1.00 & 1.00 & 0.00 & 0.78 & 0.99 & 1.00 \\
\midrule
\textbf{Syntax} 
& \multicolumn{4}{c}{\textbf{Semantic Bible}} 
& \multicolumn{4}{c}{\textbf{Family}} \\
\cmidrule(lr){2-5} \cmidrule(lr){6-9}
& \textbf{d=16} & \textbf{d=32} & \textbf{d=64} & \textbf{d=128} 
& \textbf{d=2} & \textbf{d=8} & \textbf{d=16} &  \textbf{d=32}  \\
\midrule
$A$                & 0.00 & 0.00 & 0.98 & 1.00 & 0.00 & 0.00 &0.92 & 1.00  \\
$\neg A$              & 0.00 & 0.00 & 0.98 & 1.00 & 0.00 & 0.00 & 0.98 & 1.00 \\
$C\sqcap D$          & 0.50 & 0.50 & 0.99 & 1.00 & 0.28 & 0.28 & 0.92 & 1.00  \\
$C\sqcup D$           & 0.00 & 0.00 & 0.99 & 1.00 &  0.00 & 0.00 & 0.97 & 1.00  \\
$\exists r.C$           & 0.25 & 0.28 & 0.94 & 1.00 & 0.00 & 0.00 & 0.51 & 1.00  \\
$\forall r.C$             & 0.10 & 0.10 & 0.97 & 1.00 & 0.00 & 0.00 & 0.72 & 1.00 \\
$\geq n s.C$          & 0.00 & 0.13 & 0.99 & 1.00 & 0.38 & 0.38 & 0.43 & 1.00 \\
$\leq n s.C$      & 0.00 & 0.13 & 0.97 & 1.00 & 0.00 & 0.00 & 0.95 & 1.00 \\
$\{o\}$              & 0.20 & 0.20 & 0.98 & 1.00 & 0.00 & 0.00 & 0.52 & 1.00  \\
\bottomrule
\end{tabular}
}
\label{tab:top_concept_types_datasets}
\end{table}

\begin{figure}[h!tb]
    \centering
   
    \subcaptionbox*{}
        {\includegraphics[width=0.23\textwidth]{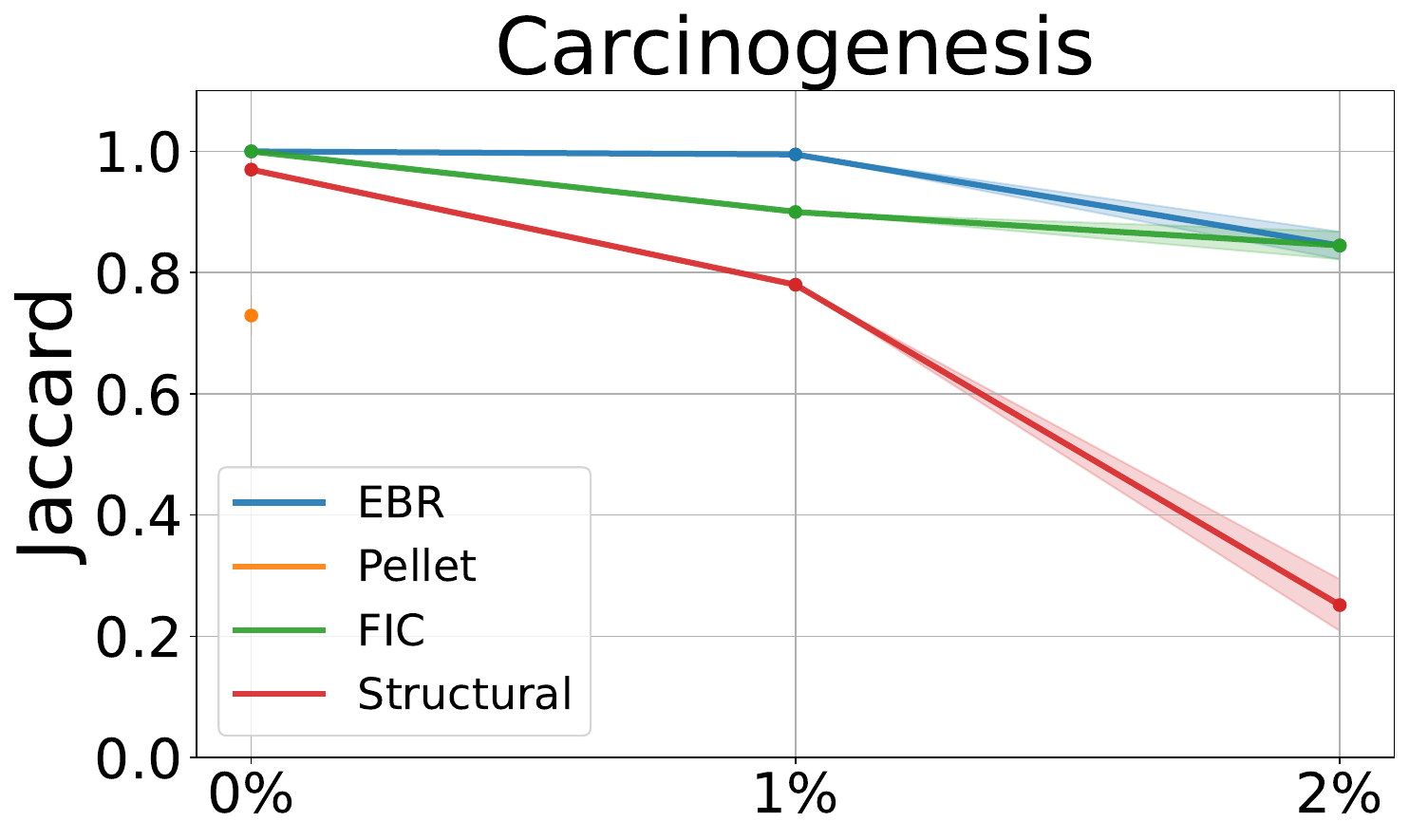}}
        \subcaptionbox*{}{\includegraphics[width=0.23\textwidth]{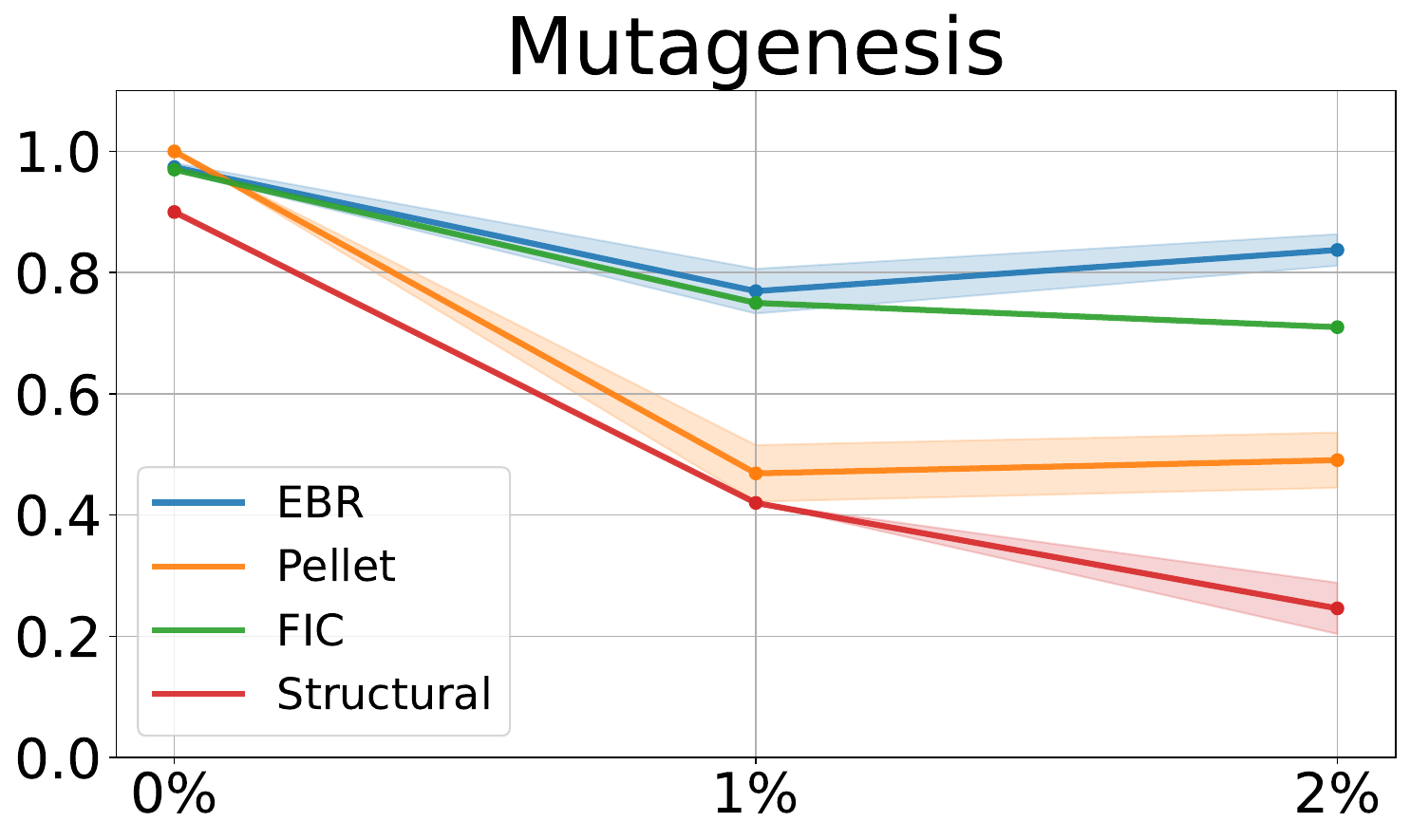}}\\
        \vspace{-.5cm}
         \subcaptionbox*{}{\includegraphics[width=0.23\textwidth]{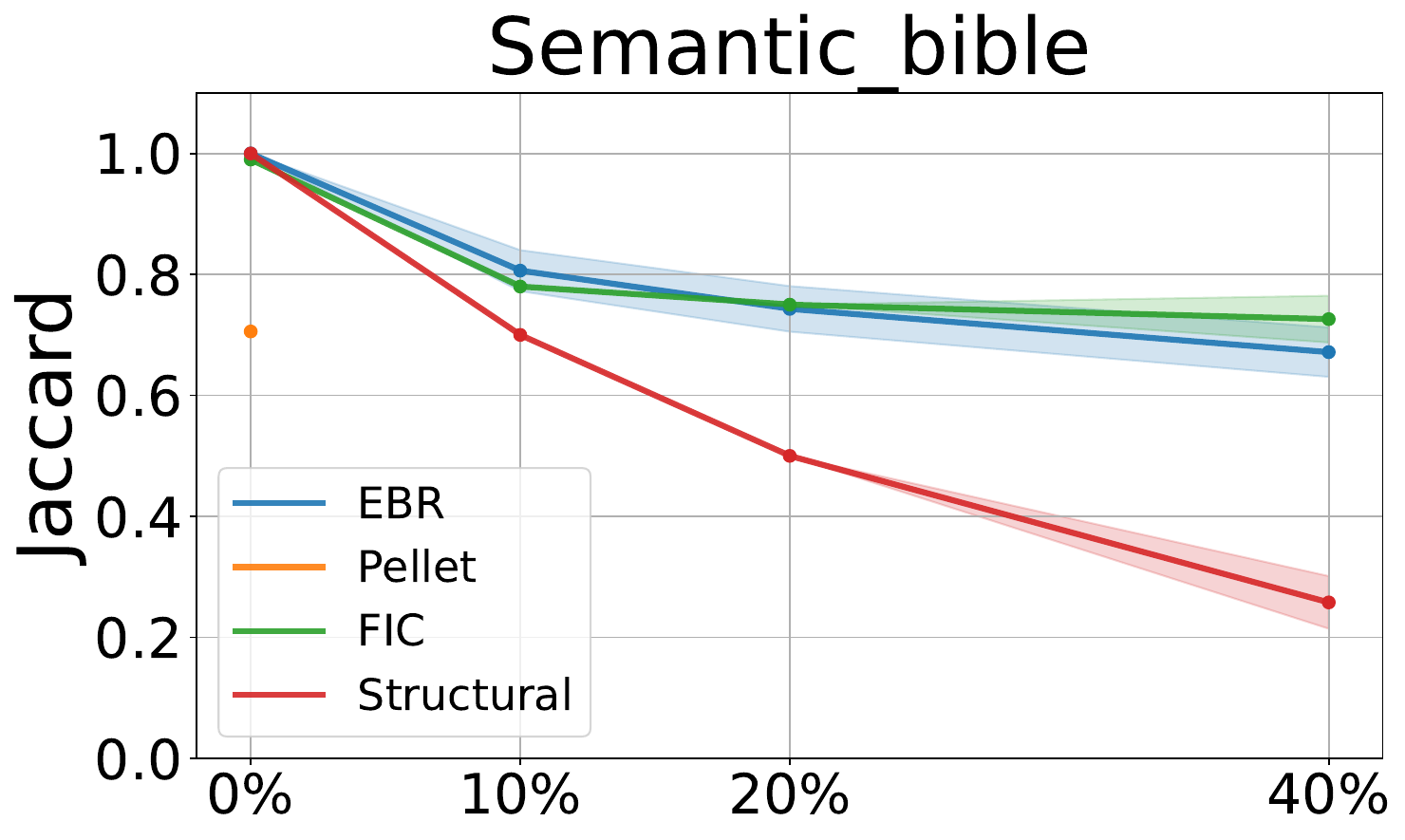}}
        \subcaptionbox*{}{\includegraphics[width=0.23\textwidth]{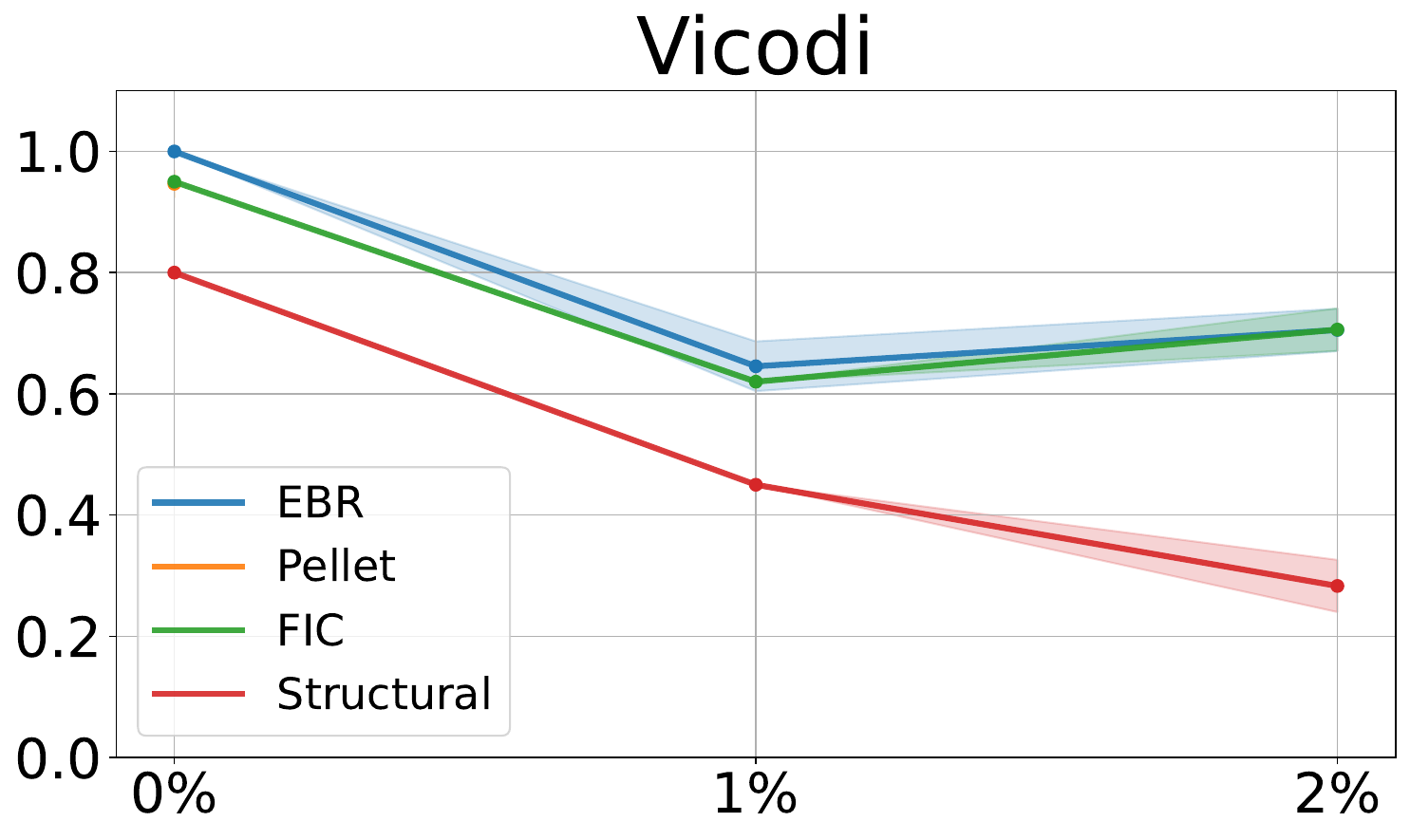}} \\
        \vspace{-.5cm}
        \subcaptionbox*{}{\includegraphics[width=0.23\textwidth,]{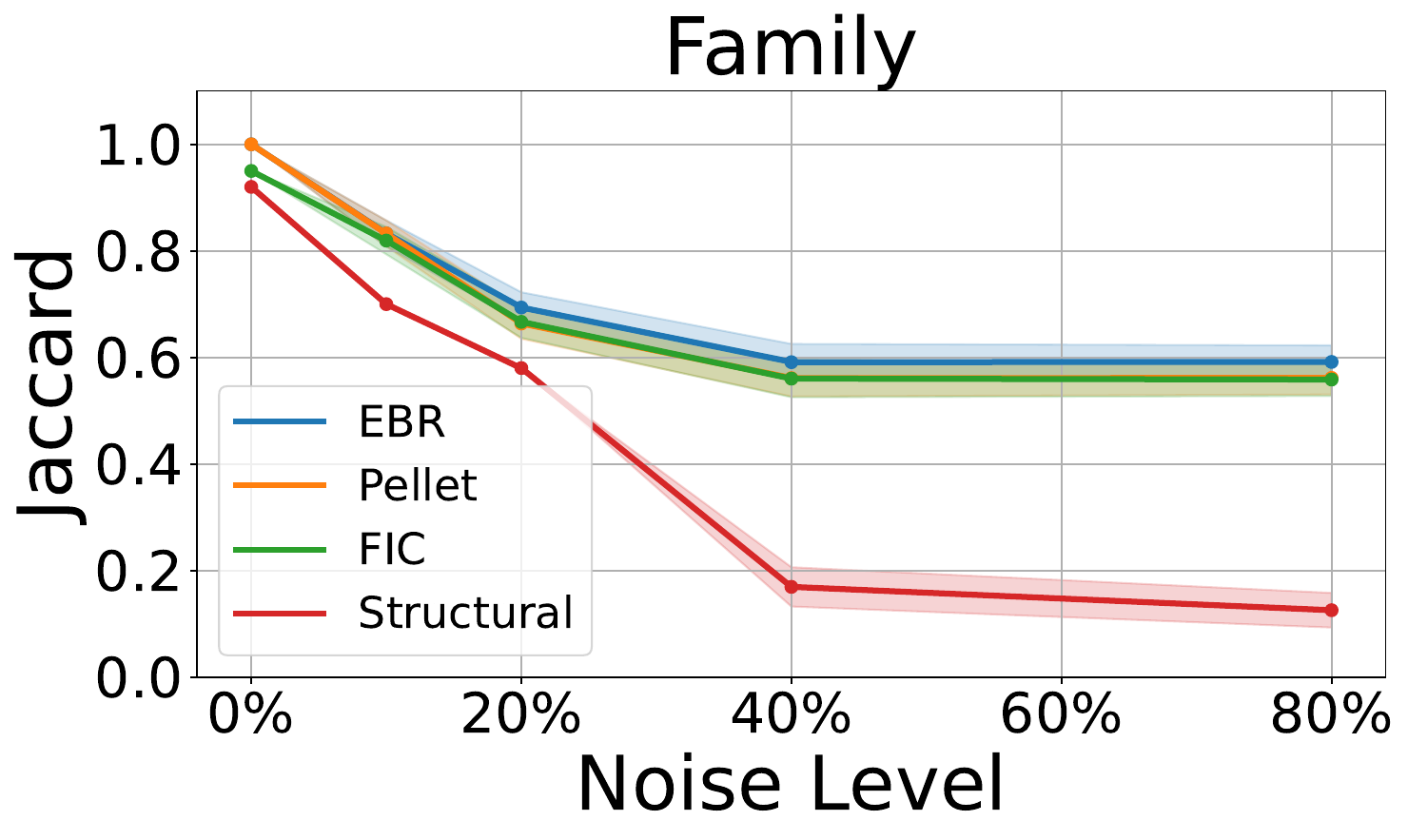}}
        \subcaptionbox*{}{\includegraphics[width=0.23\textwidth]{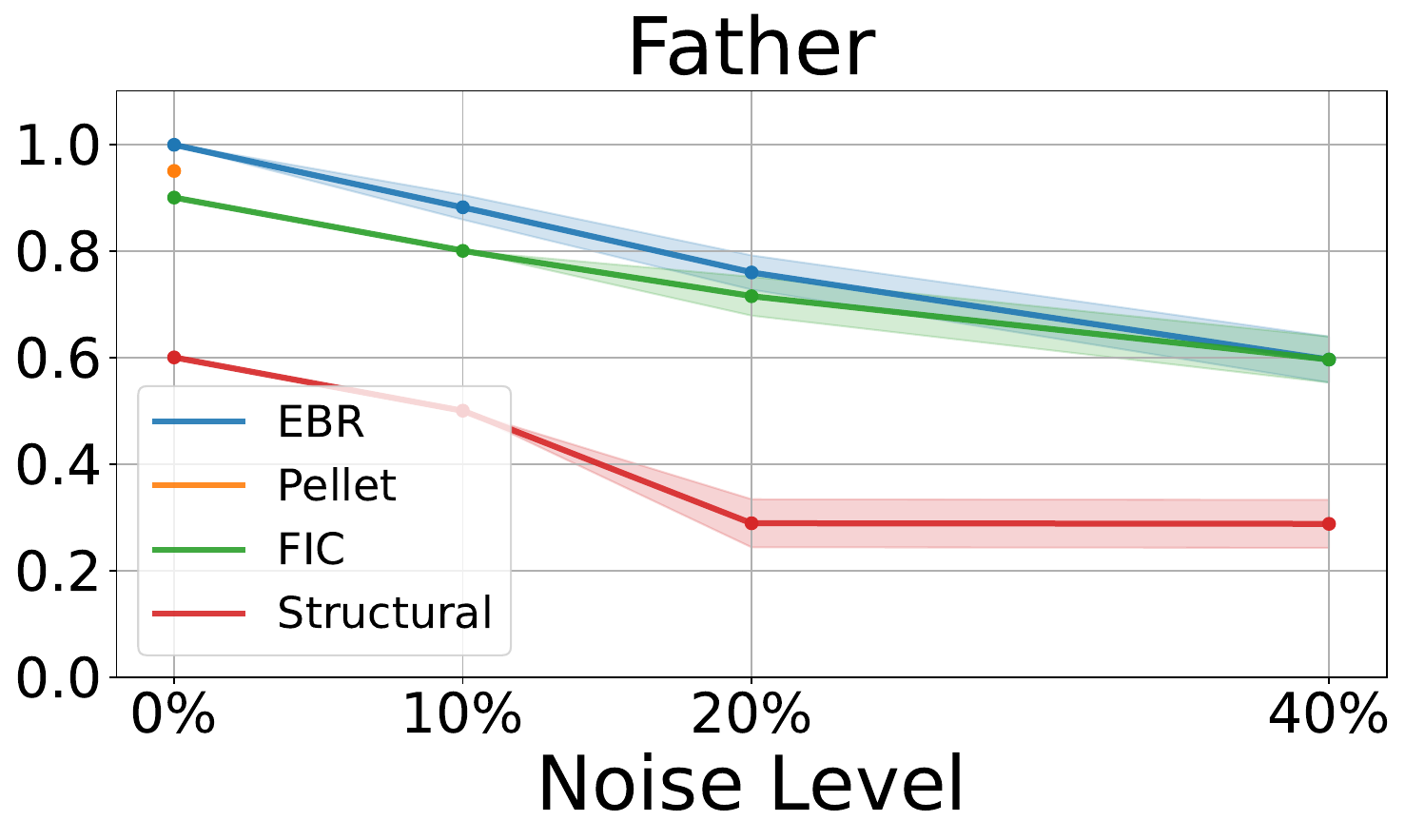}}
           \vspace{-.5cm}
  \caption{Jaccard score of each reasoner across datasets and different levels of noise}
    \label{fig:jaccard_noise_datasets}
\end{figure}
\begin{figure}[h!tb]

    \centering
    \subcaptionbox*{}{\includegraphics[width=0.23\textwidth]{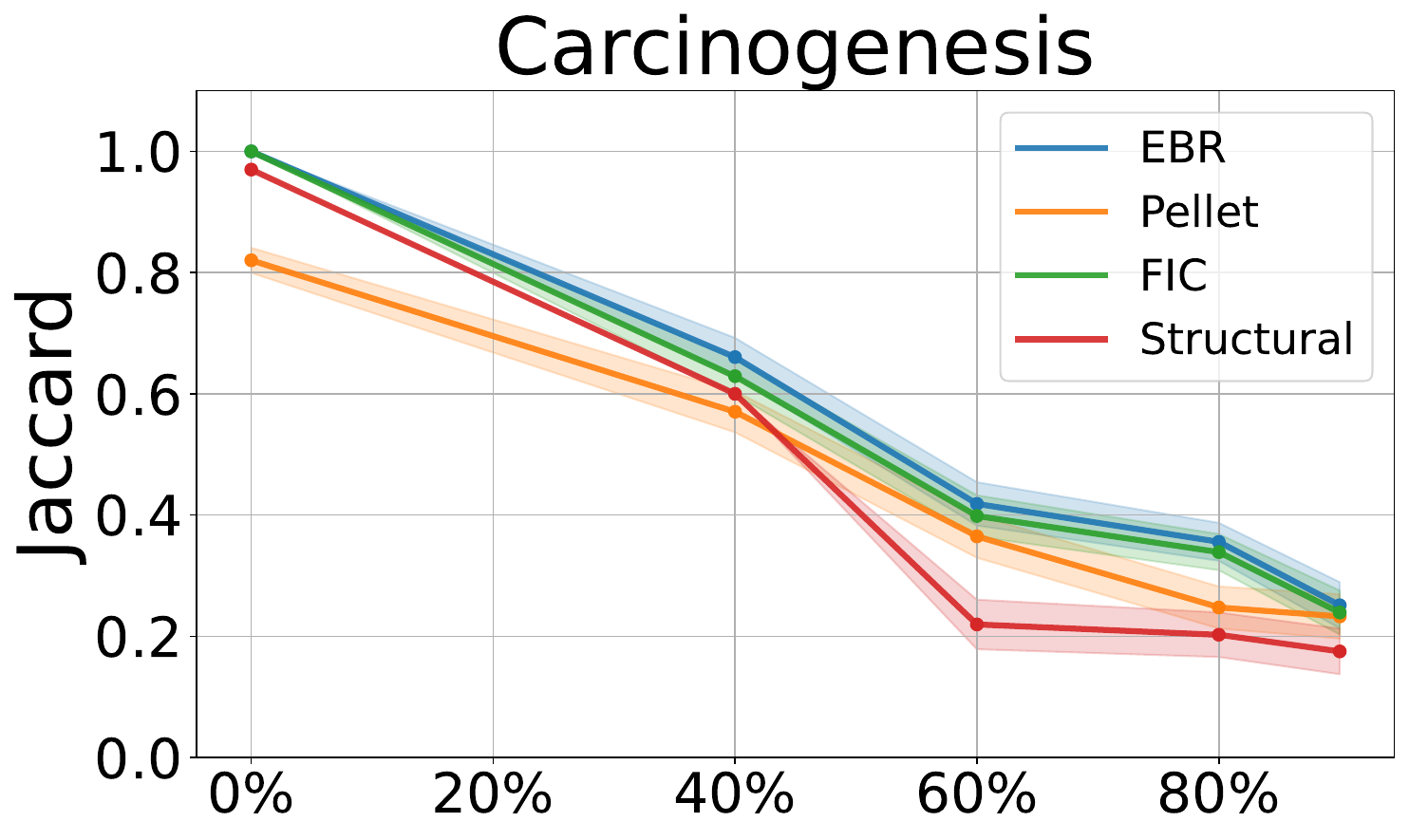}}
    \subcaptionbox*{}{\includegraphics[width=0.23\textwidth]{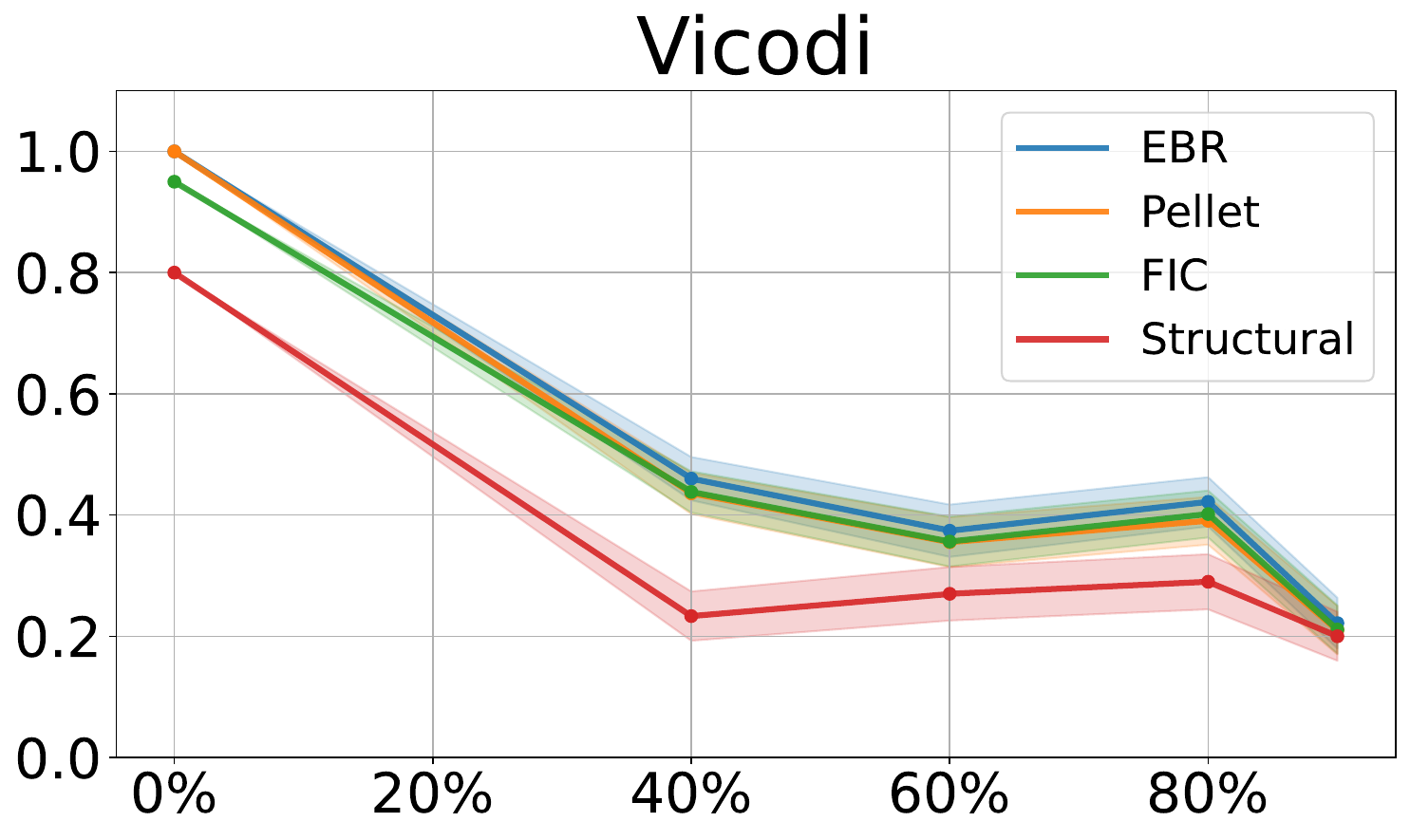}}\\
    \vspace{-.5cm}
    \subcaptionbox*{}{\includegraphics[width=0.23\textwidth]{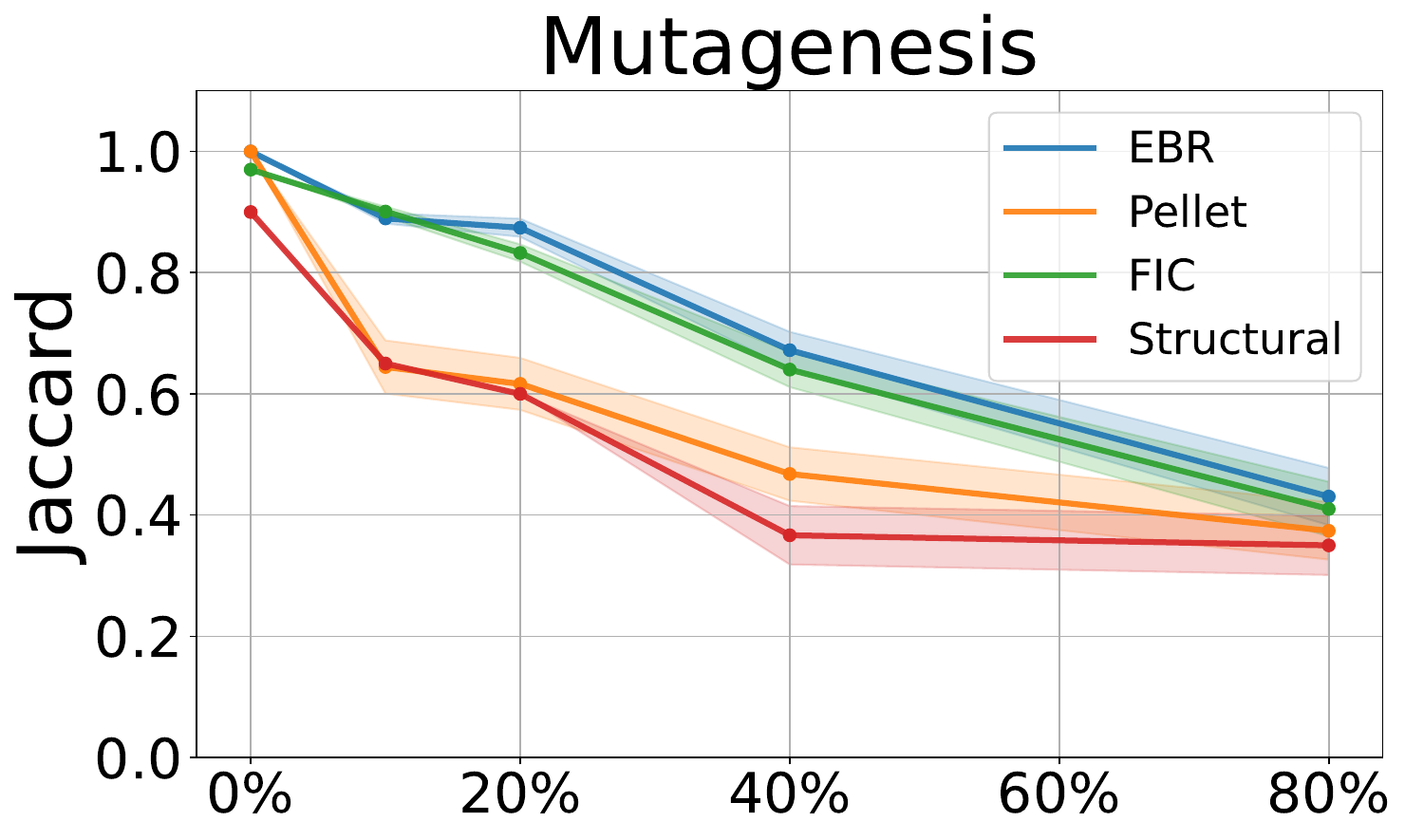}}
    \subcaptionbox*{}{\includegraphics[width=0.23\textwidth,]{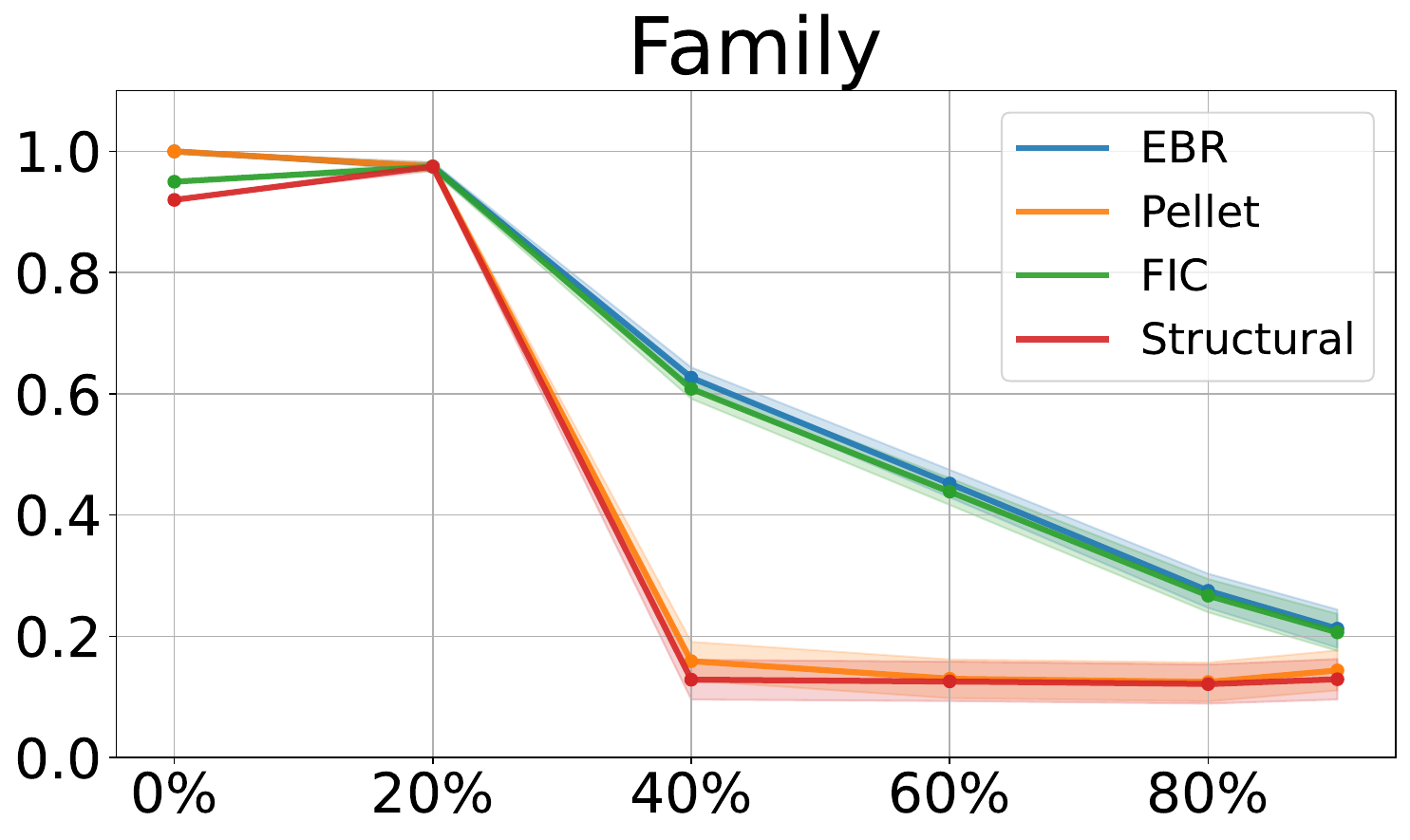}}\\
    \vspace{-.5cm}
    \subcaptionbox*{}{\includegraphics[width=0.23\textwidth]{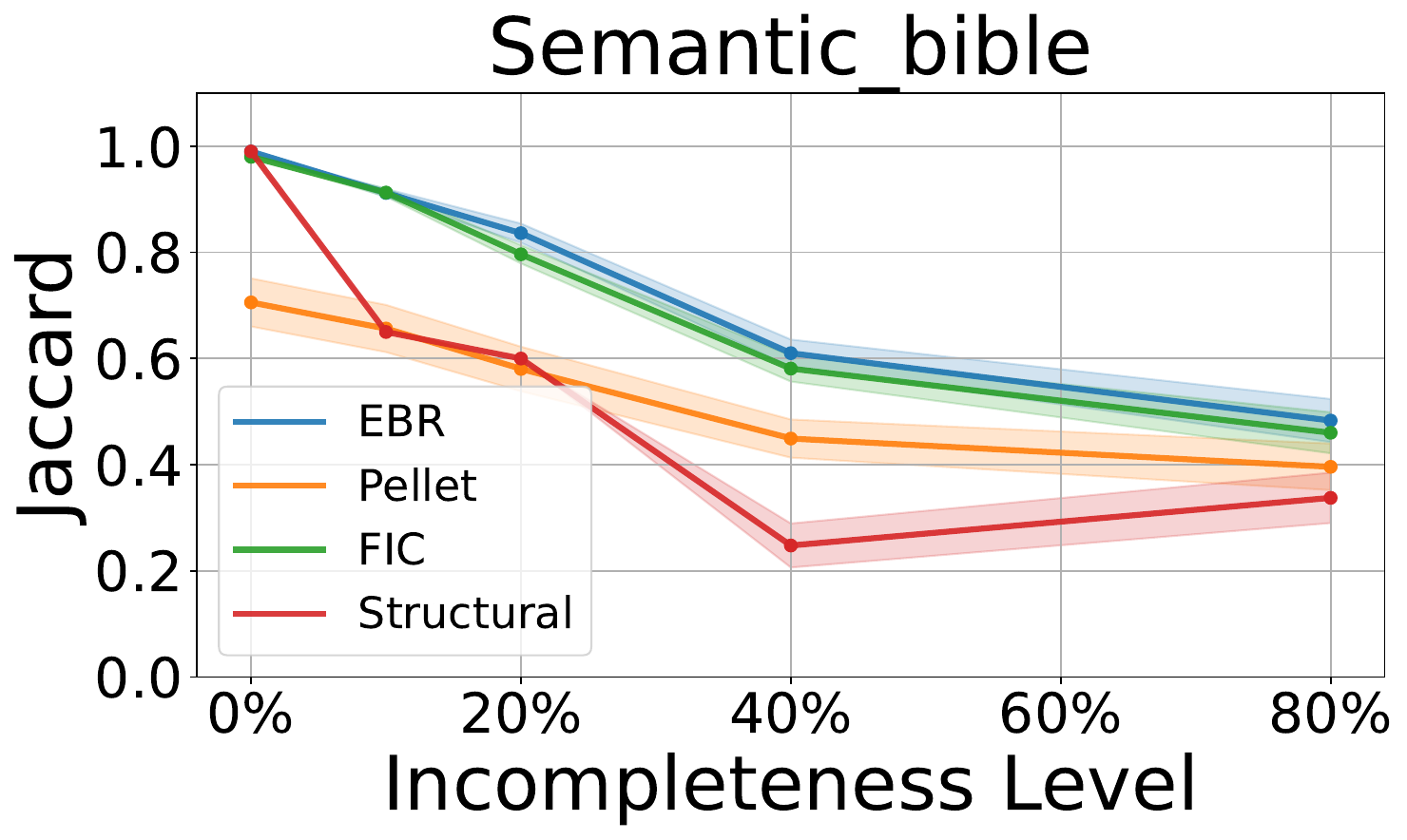}}
    \subcaptionbox*{}{\includegraphics[width=0.23\textwidth]{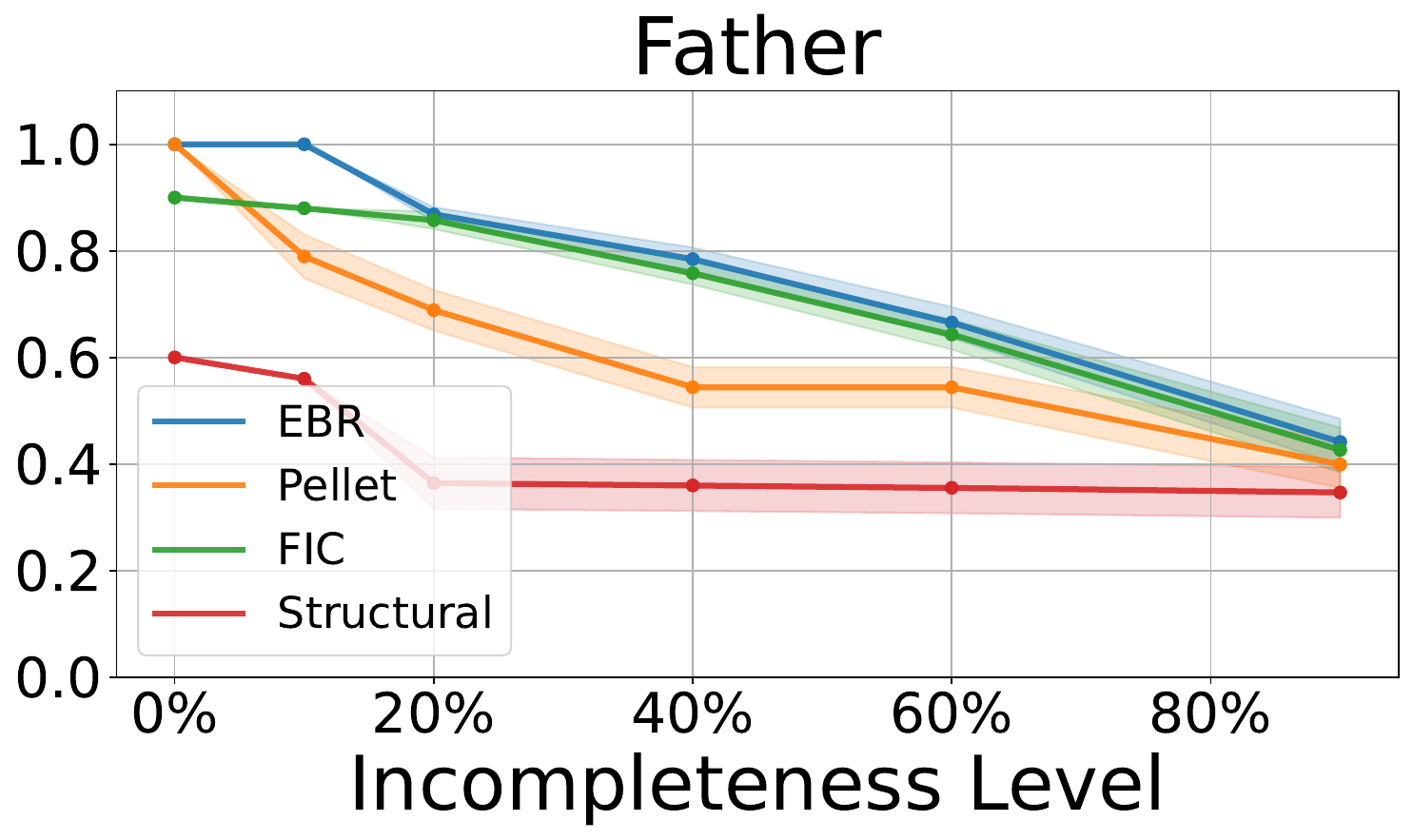}}
    \vspace{-.5cm}
 \caption{Jaccard score of each reasoner across datasets and different levels of incompleteness}
    \label{fig:jaccard_incomplete_datasets}
\end{figure}

\subsection{Retrieval on Incomplete Knowledge Bases}

Here, we present the results of $\approach$ along with state-of-the-art reasoners on all datasets.  
In Figure~\ref{fig:jaccard_incomplete_datasets}, we present the average Jaccard similarity scores with margins of error for each dataset across varying levels of incompleteness. In general, \approach exhibits higher or comparable Jaccard similarity scores relative to other reasoners in most cases, particularly at higher levels of incompleteness. However, as the level of incompleteness increases, the performance of all reasoners declines. This is expected, as higher incompleteness typically leads to poorer retrieval performance.  In this phase, removed instances are not tracked; therefore, the threshold is determined through the experiments. Therefore, given queries may not be related to removed instances, which explains why we can have an increase or the same performance for an increased level of incompleteness. 


Table~\ref{tab:avg_runtime} reports the average runtime of each reasoner across datasets for the concept retrieval task. Overall, \approach exhibits strong runtime efficiency, achieving the affordable execution times on smaller datasets such as \textit{Family} and \textit{Father}, and competitive performance on larger ones like \textit{Carcinogenesis} and \textit{Vicodi}. While symbolic reasoners such as HermiT and JFact show significantly higher runtimes, particularly on complex ontologies \approach maintains a consistent balance between speed and scalability.

\begin{table}[htb]
\centering
\setlength{\tabcolsep}{2.3pt}
\caption{Average runtime (in seconds) of each reasoner across datasets on concept retrieval (\approach runtime does not take into account training time).}
\label{tab:avg_runtime}
\resizebox{\columnwidth}{!}{%
\begin{tabular}{lcccccc}
\toprule
\textbf{Reasoner} & \textbf{Carci.} & \textbf{Muta.} & \textbf{Vicodi} & \textbf{Semantic} & \textbf{Family} & \textbf{Father} \\
\midrule
EBR      & 10.3   & 8.5   & 78.1  & 8.4   & 0.1   & 0.01 \\
HermiT   & 49.5   & 16.0   & 300.2  & 100.  & 1.0  & 0.1  \\
Pellet   & 10.1   & 2.5   & 80.2    & 6.2    & 1.5   & 0.01 \\
JFact    & 12.3   & 2.2   & 100.   & 20.9   & 4.5   & 0.04 \\
Openllet & 8.1   & 0.4   & 70.4    & 5.5   & 1.3   & 0.02 \\
FIC      & 5.0 & 0.1 & 50.5 & 5.5 & 0.05 & 0.01 \\
Structural& 2.4 & 0.1 & 25.5 & 3.6 & 0.07 & 0.01\\
\bottomrule
\end{tabular}}
\end{table}


\subsection{Retrieval on Noisy Knowledge Bases}

\begin{table}[htb]
\centering
\caption{Average F1 scores with eventual standard deviation for each concept learner across different reasoners on inconsistent datasets. The * denotes missing value due to inconsistencies.}
\label{tab:CEL_complete_inconsistent_datasets}
\resizebox{\columnwidth}{!}{%
\begin{tabular}{lcccc}
\toprule
\multicolumn{5}{c}{\textbf{Carcinogenesis (10\% Inconsistency)}} \\ 
\cmidrule(lr){2-5}&
 \textbf{Pellet}  & \textbf{EBR} & \textbf{Structural} & \textbf{FIC} \\
 \midrule
 \textbf{OCEL}  & * & $\textbf{1.00}$ &  $0.66$ & $0.93$ \\
 \textbf{CELOE} & * & $\textbf{1.00}$ &  $0.87$ & $0.93$ \\
 \textbf{EVO}   & * & $\textbf{1.00}$ & $0.77\pm 0.03$ & $0.99$ \\
 \textbf{CLIP}  & * & $\textbf{1.00}$ &  $0.87$ & $0.93$ \\

\toprule
\multicolumn{5}{c}{\textbf{Vicodi (10\% Inconsistency)}} \\ 
\cmidrule(lr){2-5}&
 \textbf{Pellet} & \textbf{EBR} & \textbf{Structural} & \textbf{FIC} \\
 \midrule
 \textbf{OCEL}  & * & $0.36 \pm 0.002$ & $0.29$ & $\textbf{0.49}$ \\
 \textbf{CELOE} & * & $\textbf{0.55} \pm \textbf{0.02}$ & $0.29$ & $0.49$ \\
 \textbf{EVO}   & * & $0.36 \pm 0.002$ &  $0.29$ & $\textbf{0.49}\pm \textbf{0.15}$ \\
 \textbf{CLIP}  & * & $\textbf{0.55} \pm \textbf{0.02}$ &  $0.29$ & $0.49$ \\
\bottomrule
\end{tabular}}

\vspace{1mm}
\begin{minipage}{\linewidth}
\end{minipage}
\end{table}

The general performance of all reasoners on noisy knowledge bases is illustrated in Figure~\ref{fig:jaccard_noise_datasets}. In the \textit{Carcinogenesis}, \textit{Mutagenesis}, and \textit{Vicodi} datasets, we injected $1\%$ and $2\%$ noise, corresponding to at least 1,000 corrupted axioms per dataset. For smaller datasets, the noise ratios ranged between $10\%$ and $80\%$. 

As shown in Figure~\ref{fig:jaccard_noise_datasets}, symbolic reasoners do not operate on most noisy datasets. Symbolic reasoners often fail on noisy datasets due to contradictions introduced by injected noise, which make knowledge bases inconsistent. In contrast, \approach, along with FIC and the Structural reasoner, can handle such inconsistencies. \approach shows strong robustness, achieving consistently high Jaccard scores and outperforming other reasoners on most datasets, except for a slight drop against FIC on the \textit{Semantic\_Bible} dataset. It also performs best on datasets without contradictions, like \textit{Mutagenesis} and \textit{Family}.

We further assessed the effectiveness of \approach when integrated into concept learning frameworks on inconsistent knowledge bases. The results in Table~\ref{tab:CEL_complete_inconsistent_datasets} report the average F1-scores (and standard deviations where applicable) for several concept learners combined with different reasoning backends. \approach maintained stable and high predictive performance across all learners. On the \textit{Carcinogenesis} dataset, EBR achieved perfect F1-scores ($1.00$) across all learners, surpassing both FIC and Structural. On the more complex \textit{Vicodi} dataset, EBR remained competitive with FIC, consistently outperforming the Structural reasoner. 
These results highlight that \approach not only supports reasoning under inconsistencies but also enables reliable concept learning in scenarios where classical symbolic reasoners cannot operate.

\section{Conclusion}

We introduced \approach, an embedding-based reasoner that leverages link prediction on knowledge graph embeddings to perform robust reasoning on inconsistent and incomplete Knowledge bases under the expressive $\mathcal{SHOIQ}$ Description Logic. 
Our experiments demonstrate that \approach significantly outperforms traditional symbolic reasoners, which do not operate with inconsistencies. 
Our experiments show that with missing assertions in a given knowledge base, \approach,  using link prediction, can still recover the full set of individuals, which is not possible for any of the traditional reasoners.

\begin{acks}
This work has received funding from the European Union's Horizon Europe research and innovation programme under grant agreements No.101070305 and No.101073307 (Marie MarieSk\l{}odowska-Curie).  It has also been supported by the project WHALE (LFN 1-04), funded under the Lamarr Fellow Network programme by the Ministry of Culture and Science of North Rhine-Westphalia (MKW NRW). In addition, this work was supported by the German Federal Ministry of Research, Technology and Space (BMFTR) within the project KI-OWL under grant No. 01IS24057B.
\end{acks}

%
\bibliographystyle{ACM-Reference-Format}
\bibliography{main}


\begin{thebibliography}{36}


\ifx \showCODEN    \undefined \def \showCODEN     #1{\unskip}     \fi
\ifx \showISBNx    \undefined \def \showISBNx     #1{\unskip}     \fi
\ifx \showISBNxiii \undefined \def \showISBNxiii  #1{\unskip}     \fi
\ifx \showISSN     \undefined \def \showISSN      #1{\unskip}     \fi
\ifx \showLCCN     \undefined \def \showLCCN      #1{\unskip}     \fi
\ifx \shownote     \undefined \def \shownote      #1{#1}          \fi
\ifx \showarticletitle \undefined \def \showarticletitle #1{#1}   \fi
\ifx \showURL      \undefined \def \showURL       {\relax}        \fi
\providecommand\bibfield[2]{#2}
\providecommand\bibinfo[2]{#2}
\providecommand\natexlab[1]{#1}
\providecommand\showeprint[2][]{arXiv:#2}

\bibitem[Baader(2003)]%
        {baader2003description}
\bibfield{author}{\bibinfo{person}{Franz Baader}.}
  \bibinfo{year}{2003}\natexlab{}.
\newblock \bibinfo{booktitle}{\emph{The description logic handbook: Theory,
  implementation and applications}}.
\newblock \bibinfo{publisher}{Cambridge university press}.
\newblock


\bibitem[Bai et~al\mbox{.}(2023)]%
        {bai2023answering}
\bibfield{author}{\bibinfo{person}{Yushi Bai}, \bibinfo{person}{Xin Lv},
  \bibinfo{person}{Juanzi Li}, {and} \bibinfo{person}{Lei Hou}.}
  \bibinfo{year}{2023}\natexlab{}.
\newblock \showarticletitle{Answering complex logical queries on knowledge
  graphs via query computation tree optimization}. In
  \bibinfo{booktitle}{\emph{International Conference on Machine Learning}}.
  PMLR, \bibinfo{pages}{1472--1491}.
\newblock


\bibitem[Bonner et~al\mbox{.}(2022)]%
        {bonner2022understanding}
\bibfield{author}{\bibinfo{person}{Stephen Bonner}, \bibinfo{person}{Ian~P
  Barrett}, \bibinfo{person}{Cheng Ye}, \bibinfo{person}{Rowan Swiers},
  \bibinfo{person}{Ola Engkvist}, \bibinfo{person}{Charles~Tapley Hoyt}, {and}
  \bibinfo{person}{William~L Hamilton}.} \bibinfo{year}{2022}\natexlab{}.
\newblock \showarticletitle{Understanding the performance of knowledge graph
  embeddings in drug discovery}.
\newblock \bibinfo{journal}{\emph{Artificial Intelligence in the Life
  Sciences}}  \bibinfo{volume}{2} (\bibinfo{year}{2022}),
  \bibinfo{pages}{100036}.
\newblock


\bibitem[Bordes et~al\mbox{.}(2013)]%
        {bordes2013translating}
\bibfield{author}{\bibinfo{person}{Antoine Bordes}, \bibinfo{person}{Nicolas
  Usunier}, \bibinfo{person}{Alberto Garcia-Duran}, \bibinfo{person}{Jason
  Weston}, {and} \bibinfo{person}{Oksana Yakhnenko}.}
  \bibinfo{year}{2013}\natexlab{}.
\newblock \showarticletitle{Translating embeddings for modeling
  multi-relational data}.
\newblock \bibinfo{journal}{\emph{Advances in neural information processing
  systems}}  \bibinfo{volume}{26} (\bibinfo{year}{2013}).
\newblock


\bibitem[Brachman and Levesque(2004)]%
        {brachman2004knowledge}
\bibfield{author}{\bibinfo{person}{Ronald Brachman} {and}
  \bibinfo{person}{Hector Levesque}.} \bibinfo{year}{2004}\natexlab{}.
\newblock \bibinfo{booktitle}{\emph{Knowledge representation and reasoning}}.
\newblock \bibinfo{publisher}{Elsevier}.
\newblock


\bibitem[B{\"u}hmann et~al\mbox{.}(2016)]%
        {buhmann2016dl}
\bibfield{author}{\bibinfo{person}{Lorenz B{\"u}hmann}, \bibinfo{person}{Jens
  Lehmann}, {and} \bibinfo{person}{Patrick Westphal}.}
  \bibinfo{year}{2016}\natexlab{}.
\newblock \showarticletitle{DL-Learner-A framework for inductive learning on
  the Semantic Web}.
\newblock \bibinfo{journal}{\emph{Journal of Web Semantics}}
  \bibinfo{volume}{39} (\bibinfo{year}{2016}), \bibinfo{pages}{15--24}.
\newblock


\bibitem[Cao et~al\mbox{.}(2021)]%
        {cao2021dual}
\bibfield{author}{\bibinfo{person}{Zongsheng Cao}, \bibinfo{person}{Qianqian
  Xu}, \bibinfo{person}{Zhiyong Yang}, \bibinfo{person}{Xiaochun Cao}, {and}
  \bibinfo{person}{Qingming Huang}.} \bibinfo{year}{2021}\natexlab{}.
\newblock \showarticletitle{Dual quaternion knowledge graph embeddings}. In
  \bibinfo{booktitle}{\emph{Proceedings of the AAAI conference on artificial
  intelligence}}, Vol.~\bibinfo{volume}{35}. \bibinfo{pages}{6894--6902}.
\newblock


\bibitem[Choudhary et~al\mbox{.}(2021)]%
        {choudhary2021self}
\bibfield{author}{\bibinfo{person}{Nurendra Choudhary}, \bibinfo{person}{Nikhil
  Rao}, \bibinfo{person}{Sumeet Katariya}, \bibinfo{person}{Karthik Subbian},
  {and} \bibinfo{person}{Chandan~K Reddy}.} \bibinfo{year}{2021}\natexlab{}.
\newblock \showarticletitle{Self-supervised hyperboloid representations from
  logical queries over knowledge graphs}. In
  \bibinfo{booktitle}{\emph{Proceedings of the Web Conference 2021}}.
  \bibinfo{pages}{1373--1384}.
\newblock


\bibitem[Dai et~al\mbox{.}(2020)]%
        {dai2020survey}
\bibfield{author}{\bibinfo{person}{Yuanfei Dai}, \bibinfo{person}{Shiping
  Wang}, \bibinfo{person}{Neal~N Xiong}, {and} \bibinfo{person}{Wenzhong Guo}.}
  \bibinfo{year}{2020}\natexlab{}.
\newblock \showarticletitle{A survey on knowledge graph embedding: Approaches,
  applications and benchmarks}.
\newblock \bibinfo{journal}{\emph{Electronics}} \bibinfo{volume}{9},
  \bibinfo{number}{5} (\bibinfo{year}{2020}), \bibinfo{pages}{750}.
\newblock


\bibitem[Demir et~al\mbox{.}(2025)]%
        {JMLR:v26:24-1113}
\bibfield{author}{\bibinfo{person}{Caglar Demir}, \bibinfo{person}{Alkid Baci},
  \bibinfo{person}{N'Dah~Jean Kouagou}, \bibinfo{person}{Leonie~Nora Sieger},
  \bibinfo{person}{Stefan Heindorf}, \bibinfo{person}{Simon Bin},
  \bibinfo{person}{Lukas Bl{{"u}}baum}, \bibinfo{person}{Alexander Bigerl},
  {and} \bibinfo{person}{Axel-Cyrille~Ngonga Ngomo}.}
  \bibinfo{year}{2025}\natexlab{}.
\newblock \showarticletitle{Ontolearn---A Framework for Large-scale OWL Class
  Expression Learning in Python}.
\newblock \bibinfo{journal}{\emph{Journal of Machine Learning Research}}
  \bibinfo{volume}{26}, \bibinfo{number}{63} (\bibinfo{year}{2025}),
  \bibinfo{pages}{1--6}.
\newblock
\urldef\tempurl%
\url{http://jmlr.org/papers/v26/24-1113.html}
\showURL{%
\tempurl}


\bibitem[Demir and Ngomo(2023)]%
        {demir2023neuro}
\bibfield{author}{\bibinfo{person}{Caglar Demir} {and}
  \bibinfo{person}{Axel-Cyrille~Ngonga Ngomo}.}
  \bibinfo{year}{2023}\natexlab{}.
\newblock \showarticletitle{Neuro-Symbolic Class Expression Learning.}. In
  \bibinfo{booktitle}{\emph{IJCAI}}. \bibinfo{pages}{3624--3632}.
\newblock


\bibitem[Demir and Ngonga~Ngomo(2023)]%
        {demir2023clifford}
\bibfield{author}{\bibinfo{person}{Caglar Demir} {and}
  \bibinfo{person}{Axel-Cyrille Ngonga~Ngomo}.}
  \bibinfo{year}{2023}\natexlab{}.
\newblock \showarticletitle{Clifford Embeddings--A Generalized Approach for
  Embedding in Normed Algebras}. In \bibinfo{booktitle}{\emph{Joint European
  Conference on Machine Learning and Knowledge Discovery in Databases}}.
  Springer, \bibinfo{pages}{567--582}.
\newblock


\bibitem[Demir et~al\mbox{.}(2023)]%
        {demir2023litcqd}
\bibfield{author}{\bibinfo{person}{Caglar Demir}, \bibinfo{person}{Michel
  Wiebesiek}, \bibinfo{person}{Renzhong Lu}, \bibinfo{person}{Axel-Cyrille
  Ngonga~Ngomo}, {and} \bibinfo{person}{Stefan Heindorf}.}
  \bibinfo{year}{2023}\natexlab{}.
\newblock \showarticletitle{LitCQD: Multi-hop Reasoning in Incomplete Knowledge
  Graphs with Numeric Literals}. In \bibinfo{booktitle}{\emph{Joint European
  Conference on Machine Learning and Knowledge Discovery in Databases}}.
  Springer, \bibinfo{pages}{617--633}.
\newblock


\bibitem[Dettmers et~al\mbox{.}(2018)]%
        {dettmers2018convolutional}
\bibfield{author}{\bibinfo{person}{Tim Dettmers}, \bibinfo{person}{Pasquale
  Minervini}, \bibinfo{person}{Pontus Stenetorp}, {and}
  \bibinfo{person}{Sebastian Riedel}.} \bibinfo{year}{2018}\natexlab{}.
\newblock \showarticletitle{Convolutional 2D Knowledge Graph Embeddings}. In
  \bibinfo{booktitle}{\emph{{AAAI}}}. \bibinfo{publisher}{{AAAI} Press},
  \bibinfo{pages}{1811--1818}.
\newblock


\bibitem[Glimm et~al\mbox{.}(2014)]%
        {glimm2014hermit}
\bibfield{author}{\bibinfo{person}{Birte Glimm}, \bibinfo{person}{Ian
  Horrocks}, \bibinfo{person}{Boris Motik}, \bibinfo{person}{Giorgos Stoilos},
  {and} \bibinfo{person}{Zhe Wang}.} \bibinfo{year}{2014}\natexlab{}.
\newblock \showarticletitle{HermiT: an OWL 2 reasoner}.
\newblock \bibinfo{journal}{\emph{Journal of automated reasoning}}
  \bibinfo{volume}{53} (\bibinfo{year}{2014}), \bibinfo{pages}{245--269}.
\newblock


\bibitem[Hamilton et~al\mbox{.}(2018)]%
        {hamilton2018embedding}
\bibfield{author}{\bibinfo{person}{Will Hamilton}, \bibinfo{person}{Payal
  Bajaj}, \bibinfo{person}{Marinka Zitnik}, \bibinfo{person}{Dan Jurafsky},
  {and} \bibinfo{person}{Jure Leskovec}.} \bibinfo{year}{2018}\natexlab{}.
\newblock \showarticletitle{Embedding logical queries on knowledge graphs}.
\newblock \bibinfo{journal}{\emph{Advances in neural information processing
  systems}}  \bibinfo{volume}{31} (\bibinfo{year}{2018}).
\newblock


\bibitem[Heindorf et~al\mbox{.}(2022)]%
        {heindorf2022evolearner}
\bibfield{author}{\bibinfo{person}{Stefan Heindorf}, \bibinfo{person}{Lukas
  Bl{\"u}baum}, \bibinfo{person}{Nick D{\"u}sterhus}, \bibinfo{person}{Till
  Werner}, \bibinfo{person}{Varun~Nandkumar Golani}, \bibinfo{person}{Caglar
  Demir}, {and} \bibinfo{person}{Axel-Cyrille Ngonga~Ngomo}.}
  \bibinfo{year}{2022}\natexlab{}.
\newblock \showarticletitle{Evolearner: Learning description logics with
  evolutionary algorithms}. In \bibinfo{booktitle}{\emph{Proceedings of the ACM
  Web Conference 2022}}. \bibinfo{pages}{818--828}.
\newblock


\bibitem[Hitzler et~al\mbox{.}(2009)]%
        {hitzler2009foundations}
\bibfield{author}{\bibinfo{person}{Pascal Hitzler}, \bibinfo{person}{Markus
  Krotzsch}, {and} \bibinfo{person}{Sebastian Rudolph}.}
  \bibinfo{year}{2009}\natexlab{}.
\newblock \bibinfo{booktitle}{\emph{Foundations of semantic web technologies}}.
\newblock \bibinfo{publisher}{Chapman and Hall/CRC}.
\newblock


\bibitem[Horrocks et~al\mbox{.}(2003)]%
        {horrocks2003shiq}
\bibfield{author}{\bibinfo{person}{Ian Horrocks}, \bibinfo{person}{Peter~F
  Patel-Schneider}, {and} \bibinfo{person}{Frank Van~Harmelen}.}
  \bibinfo{year}{2003}\natexlab{}.
\newblock \showarticletitle{From SHIQ and RDF to OWL: The making of a web
  ontology language}.
\newblock \bibinfo{journal}{\emph{Journal of web semantics}}
  \bibinfo{volume}{1}, \bibinfo{number}{1} (\bibinfo{year}{2003}),
  \bibinfo{pages}{7--26}.
\newblock


\bibitem[Kazakov(2008)]%
        {Kazakov08}
\bibfield{author}{\bibinfo{person}{Yevgeny Kazakov}.}
  \bibinfo{year}{2008}\natexlab{}.
\newblock \showarticletitle{{RIQ} and {SROIQ} Are Harder than {SHOIQ}}. In
  \bibinfo{booktitle}{\emph{Principles of Knowledge Representation and
  Reasoning: Proceedings of the Eleventh International Conference, {KR} 2008,
  Sydney, Australia, September 16-19, 2008}},
  \bibfield{editor}{\bibinfo{person}{Gerhard Brewka} {and}
  \bibinfo{person}{J{\'{e}}r{\^{o}}me Lang}} (Eds.). \bibinfo{publisher}{{AAAI}
  Press}, \bibinfo{pages}{274--284}.
\newblock
\urldef\tempurl%
\url{http://www.aaai.org/Library/KR/2008/kr08-027.php}
\showURL{%
\tempurl}


\bibitem[Keet(2018)]%
        {keet2018introduction}
\bibfield{author}{\bibinfo{person}{C~Maria Keet}.}
  \bibinfo{year}{2018}\natexlab{}.
\newblock \showarticletitle{An introduction to ontology engineering}.
\newblock  (\bibinfo{year}{2018}).
\newblock


\bibitem[Kouagou et~al\mbox{.}(2022)]%
        {kouagou2022learning}
\bibfield{author}{\bibinfo{person}{N'Dah~Jean Kouagou}, \bibinfo{person}{Stefan
  Heindorf}, \bibinfo{person}{Caglar Demir}, {and}
  \bibinfo{person}{Axel-Cyrille~Ngonga Ngomo}.}
  \bibinfo{year}{2022}\natexlab{}.
\newblock \showarticletitle{Learning concept lengths accelerates concept
  learning in ALC}. In \bibinfo{booktitle}{\emph{European Semantic Web
  Conference}}. Springer, \bibinfo{pages}{236--252}.
\newblock


\bibitem[Lehmann and Hitzler(2010)]%
        {lehmann2010concept}
\bibfield{author}{\bibinfo{person}{Jens Lehmann} {and} \bibinfo{person}{Pascal
  Hitzler}.} \bibinfo{year}{2010}\natexlab{}.
\newblock \showarticletitle{Concept learning in description logics using
  refinement operators}.
\newblock \bibinfo{journal}{\emph{Machine Learning}}  \bibinfo{volume}{78}
  (\bibinfo{year}{2010}), \bibinfo{pages}{203--250}.
\newblock


\bibitem[Nickel et~al\mbox{.}(2015)]%
        {nickel2015review}
\bibfield{author}{\bibinfo{person}{Maximilian Nickel}, \bibinfo{person}{Kevin
  Murphy}, \bibinfo{person}{Volker Tresp}, {and} \bibinfo{person}{Evgeniy
  Gabrilovich}.} \bibinfo{year}{2015}\natexlab{}.
\newblock \showarticletitle{A review of relational machine learning for
  knowledge graphs}.
\newblock \bibinfo{journal}{\emph{Proc. IEEE}} \bibinfo{volume}{104},
  \bibinfo{number}{1} (\bibinfo{year}{2015}), \bibinfo{pages}{11--33}.
\newblock


\bibitem[Qian et~al\mbox{.}(2021)]%
        {qian2021understanding}
\bibfield{author}{\bibinfo{person}{Jing Qian}, \bibinfo{person}{Gangmin Li},
  \bibinfo{person}{Katie Atkinson}, {and} \bibinfo{person}{Yong Yue}.}
  \bibinfo{year}{2021}\natexlab{}.
\newblock \showarticletitle{Understanding negative sampling in knowledge graph
  embedding}.
\newblock  (\bibinfo{year}{2021}).
\newblock


\bibitem[Ren and Leskovec(2020)]%
        {ren2020beta}
\bibfield{author}{\bibinfo{person}{Hongyu Ren} {and} \bibinfo{person}{Jure
  Leskovec}.} \bibinfo{year}{2020}\natexlab{}.
\newblock \showarticletitle{Beta embeddings for multi-hop logical reasoning in
  knowledge graphs}.
\newblock \bibinfo{journal}{\emph{Advances in Neural Information Processing
  Systems}}  \bibinfo{volume}{33} (\bibinfo{year}{2020}),
  \bibinfo{pages}{19716--19726}.
\newblock


\bibitem[Ruffinelli et~al\mbox{.}(2020)]%
        {Rufinelli2020You}
\bibfield{author}{\bibinfo{person}{Daniel Ruffinelli}, \bibinfo{person}{Samuel
  Broscheit}, {and} \bibinfo{person}{Rainer Gemulla}.}
  \bibinfo{year}{2020}\natexlab{}.
\newblock \showarticletitle{You {CAN} Teach an Old Dog New Tricks! On Training
  Knowledge Graph Embeddings}. In \bibinfo{booktitle}{\emph{{ICLR}}}.
  \bibinfo{publisher}{OpenReview.net}.
\newblock


\bibitem[Sattler et~al\mbox{.}(2014)]%
        {sattler2014does}
\bibfield{author}{\bibinfo{person}{Uli Sattler}, \bibinfo{person}{Robert
  Stevens}, {and} \bibinfo{person}{Phillip Lord}.}
  \bibinfo{year}{2014}\natexlab{}.
\newblock \showarticletitle{How does a reasoner work?}
\newblock \bibinfo{journal}{\emph{Ontogenesis}} (\bibinfo{year}{2014}).
\newblock


\bibitem[Singh et~al\mbox{.}(2020)]%
        {Singh2020OWL2Bench}
\bibfield{author}{\bibinfo{person}{Gunjan Singh}, \bibinfo{person}{Sumit
  Bhatia}, {and} \bibinfo{person}{Raghava Mutharaju}.}
  \bibinfo{year}{2020}\natexlab{}.
\newblock \showarticletitle{OWL2Bench: {A} Benchmark for {OWL} 2 Reasoners}. In
  \bibinfo{booktitle}{\emph{{ISWC} {(2)}}} \emph{(\bibinfo{series}{Lecture
  Notes in Computer Science}, Vol.~\bibinfo{volume}{12507})}.
  \bibinfo{publisher}{Springer}, \bibinfo{pages}{81--96}.
\newblock


\bibitem[Sirin et~al\mbox{.}(2007)]%
        {sirin2007pellet}
\bibfield{author}{\bibinfo{person}{Evren Sirin}, \bibinfo{person}{Bijan
  Parsia}, \bibinfo{person}{Bernardo~Cuenca Grau}, \bibinfo{person}{Aditya
  Kalyanpur}, {and} \bibinfo{person}{Yarden Katz}.}
  \bibinfo{year}{2007}\natexlab{}.
\newblock \showarticletitle{Pellet: A practical owl-dl reasoner}.
\newblock \bibinfo{journal}{\emph{Journal of Web Semantics}}
  \bibinfo{volume}{5}, \bibinfo{number}{2} (\bibinfo{year}{2007}),
  \bibinfo{pages}{51--53}.
\newblock


\bibitem[Teyou et~al\mbox{.}(2024)]%
        {teyou2024embedding}
\bibfield{author}{\bibinfo{person}{Louis Mozart~Kamdem Teyou},
  \bibinfo{person}{Caglar Demir}, {and} \bibinfo{person}{Axel-Cyrille~Ngonga
  Ngomo}.} \bibinfo{year}{2024}\natexlab{}.
\newblock \showarticletitle{Embedding Knowledge Graphs in Degenerate Clifford
  Algebras}.
\newblock \bibinfo{journal}{\emph{arXiv preprint arXiv:2402.04870}}
  (\bibinfo{year}{2024}).
\newblock


\bibitem[Trouillon et~al\mbox{.}(2017)]%
        {trouillon2017knowledge}
\bibfield{author}{\bibinfo{person}{Th{\'e}o Trouillon},
  \bibinfo{person}{Christopher~R Dance}, \bibinfo{person}{{\'E}ric Gaussier},
  \bibinfo{person}{Johannes Welbl}, \bibinfo{person}{Sebastian Riedel}, {and}
  \bibinfo{person}{Guillaume Bouchard}.} \bibinfo{year}{2017}\natexlab{}.
\newblock \showarticletitle{Knowledge graph completion via complex tensor
  factorization}.
\newblock \bibinfo{journal}{\emph{Journal of Machine Learning Research}}
  \bibinfo{volume}{18}, \bibinfo{number}{130} (\bibinfo{year}{2017}),
  \bibinfo{pages}{1--38}.
\newblock


\bibitem[Tsarkov and Horrocks(2006)]%
        {tsarkov2006fact++}
\bibfield{author}{\bibinfo{person}{Dmitry Tsarkov} {and} \bibinfo{person}{Ian
  Horrocks}.} \bibinfo{year}{2006}\natexlab{}.
\newblock \showarticletitle{FaCT++ description logic reasoner: System
  description}. In \bibinfo{booktitle}{\emph{International joint conference on
  automated reasoning}}. Springer, \bibinfo{pages}{292--297}.
\newblock


\bibitem[Wang et~al\mbox{.}(2021)]%
        {wang2021survey}
\bibfield{author}{\bibinfo{person}{Meihong Wang}, \bibinfo{person}{Linling
  Qiu}, {and} \bibinfo{person}{Xiaoli Wang}.} \bibinfo{year}{2021}\natexlab{}.
\newblock \showarticletitle{A survey on knowledge graph embeddings for link
  prediction}.
\newblock \bibinfo{journal}{\emph{Symmetry}} \bibinfo{volume}{13},
  \bibinfo{number}{3} (\bibinfo{year}{2021}), \bibinfo{pages}{485}.
\newblock


\bibitem[Werner et~al\mbox{.}(2014)]%
        {werner2014using}
\bibfield{author}{\bibinfo{person}{David Werner}, \bibinfo{person}{Nuno Silva},
  \bibinfo{person}{Christophe Cruz}, {and} \bibinfo{person}{Aurelie Bertaux}.}
  \bibinfo{year}{2014}\natexlab{}.
\newblock \showarticletitle{Using DL-reasoner for hierarchical multilabel
  classification applied to economical e-news}. In
  \bibinfo{booktitle}{\emph{2014 Science and Information Conference}}. IEEE,
  \bibinfo{pages}{313--320}.
\newblock


\bibitem[Yang et~al\mbox{.}(2014)]%
        {yang2014embedding}
\bibfield{author}{\bibinfo{person}{Bishan Yang}, \bibinfo{person}{Wen-tau Yih},
  \bibinfo{person}{Xiaodong He}, \bibinfo{person}{Jianfeng Gao}, {and}
  \bibinfo{person}{Li Deng}.} \bibinfo{year}{2014}\natexlab{}.
\newblock \showarticletitle{Embedding entities and relations for learning and
  inference in knowledge bases}.
\newblock \bibinfo{journal}{\emph{arXiv preprint arXiv:1412.6575}}
  (\bibinfo{year}{2014}).
\newblock


\end{thebibliography}

\end{document}